\documentclass[journal]{IEEEtran}
\makeatletter
\def\markboth#1#2{\def\leftmark{\@IEEEcompsoconly{\sffamily}\MakeUppercase{\protect#1}}%
\def\rightmark{\@IEEEcompsoconly{\sffamily}\MakeUppercase{\protect#2}}}
\makeatother

\ifCLASSINFOpdf
\else
\fi

\usepackage[font=scriptsize]{caption}
\usepackage[font=scriptsize]{subfig}
\let\MYorigsubfloat\subfloat
\renewcommand{\subfloat}[2][\relax]{\MYorigsubfloat[]{#2}}
\usepackage[T1]{fontenc}
\usepackage[latin1]{inputenc}
\usepackage[swedish,english]{babel}

\usepackage[percent]{overpic}
\usepackage{rotating}
\usepackage{tabularx}
\usepackage{xcolor}

\newcommand{\V}{$\surd$}

\usepackage[numbers]{natbib}

\usepackage[textwidth=2cm,colorinlistoftodos]{todonotes}

\usepackage{tikz}
\usepackage{xcolor}
\usetikzlibrary{fit,backgrounds,shadows,shapes,calc,positioning,mindmap,decorations,arrows,shapes, matrix}

\usepackage{pgfplots}
\usepackage{pgfplotstable}
\pgfplotsset{ 
compat=1.3,
tick label style={font=\footnotesize}, 
label style={font=\footnotesize}, 
legend style={font=\footnotesize}
}
\tikzstyle{module} = [draw, 
  rectangle, 
  align=flush center, 
  font=\scriptsize\bfseries, 
  text width = 1.8cm, 
  minimum height = 1cm, 
  minimum width = 2cm ]
  
\tikzstyle{scene} = [draw, 
  ellipse,
  minimum height = 1 cm, 
  minimum width = 2cm,
  align=flush center, 
  font=\footnotesize, 
  text width = 1.8cm]  
  
\tikzstyle{execution} = [draw, 
  ellipse, 
  minimum height = 1 cm,
  minimum width = 2cm,
  align=flush center, 
  font=\footnotesize, 
  text width = 1.8cm]
  
\tikzstyle{db} = [draw,
  cylinder,
  shape border rotate = 90,
  minimum height = 1 cm,
  minimum width = 2cm,
  aspect = 0.1,
  align=flush center, 
  font=\scriptsize\bfseries, 
  text width = 1.8cm ]
  
\tikzstyle{label} = [ 
  align = flush center,
  text width =1cm,
  font = \tiny,
  text = violet ]
  
\tikzstyle{model} = [draw,  
  rectangle,
  align=flush center,
  rounded corners = 8pt,
  very thick,
  font=\scriptsize\bfseries,
  text width = 1.8 cm,
  minimum width = 2cm,
  minimum height = 1cm ]

\usepackage{amsmath}
\usepackage{url}
\usepackage{multirow}
\usepackage{booktabs}

\usepackage{tikz}
\usepackage{textcomp}
\usepackage{hyperref}

\usepackage[T1]{fontenc}
\usepackage[latin1]{inputenc}
\usepackage[swedish,english]{babel}
\graphicspath{{./figures/related}}

\begin{document}
%
\title{Data-Driven Grasp Synthesis - A Survey}
%
%
%

\author{Jeannette Bohg,~\IEEEmembership{Member,~IEEE,} 
Antonio Morales,~\IEEEmembership{Member,~IEEE,}
Tamim Asfour,~\IEEEmembership{Member,~IEEE,}
\\ Danica Kragic~\IEEEmembership{Member,~IEEE}
\thanks{J. Bohg is with the Autonomous Motion Department at the MPI for Intelligent
  Systems, T{\"u}bingen, Germany, e-mail:
  jbohg@tuebingen.mpg.de.}
\thanks{A. Morales is with the Robotic Intelligence Lab. at Universitat Jaume I, Castell{\'o}, 
Spain, e-mail:
  Antonio.Morales@uji.es.}
\thanks{T. Asfour is with the KIT, Karlsruhe, Germany, e-mail:
  asfour@kit.edu.}
\thanks{D. Kragic is with the Centre for Autonomous Systems, Computational Vision and Active
  Perception Lab, Royal Institute fo Technology KTH,  Stockholm, Sweden, e-mail: dank@kth.se.}
\thanks{This work has been supported by FLEXBOT (FP7-ERC-279933). }
}

%
%

\markboth{Transactions on Robotics}%
{Bohg \MakeLowercase{\textit{et al.}}: Data-Driven Grasp Synthesis - A Survey}
%


\newcommand\copyrighttext{%
  \footnotesize \textcopyright 2014 IEEE. Personal use of this material is permitted.
  Permission from IEEE must be obtained for all other uses, in any current or future
  media, including reprinting/republishing this material for advertising or promotional
  purposes, creating new collective works, for resale or redistribution to servers or
  lists, or reuse of any copyrighted component of this work in other works.
  DOI: \href{http://dx.doi.org/10.1109/TRO.2013.2289018}{<10.1109/TRO.2013.2289018>}}
\newcommand\copyrightnotice{%
\begin{tikzpicture}[remember picture,overlay]
\node[anchor=south,yshift=10pt] at (current page.south) {\fbox{\parbox{\dimexpr\textwidth-\fboxsep-\fboxrule\relax}{\copyrighttext}}};
\end{tikzpicture}%
}


\maketitle
\copyrightnotice

\begin{abstract}
We review the work on data-driven grasp synthesis and the
methodologies for sampling and ranking candidate grasps.  We
divide the approaches into three groups based on whether they
synthesize grasps for known, familiar or unknown objects.  This
structure allows us to identify common object representations and
perceptual processes that facilitate the employed data-driven grasp
synthesis technique.
In the case of known objects, we concentrate on the approaches that
are based on object recognition and pose estimation. In the case of
familiar objects, the techniques use some form of a
similarity matching to a set of previously encountered objects.
Finally, for the approaches dealing with unknown objects, the core part
is the extraction of specific features that are indicative of good
grasps. 
Our survey provides an overview of the different methodologies and
discusses open problems in the area of robot grasping. We also draw a
parallel to the classical approaches that rely on analytic
formulations.



\end{abstract}


%
\IEEEpeerreviewmaketitle



\section{Introduction}
\label{sec:introduction}
Given an object, {\em grasp synthesis\/} refers to the problem of
finding a grasp configuration that satisfies a set of criteria
relevant for the grasping task. Finding a suitable grasp among the
infinite set of candidates is a challenging problem and has been
addressed frequently in the robotics community, resulting in an
abundance of approaches.

In the recent review of \citet{Sahbani:2011}, the authors
divide the methodologies into {\em analytic\/} and {\em
empirical\/}. Following~\citet{Shimoga06}, analytic refers to methods that 
construct force-closure grasps with a multi-fingered
robotic hand that are 
{\em dexterous\/}, in {\em equilibrium\/}, {\em stable\/} and
exhibit a certain {\em dynamic behaviour\/}. Grasp synthesis is then
usually formulated as a constrained optimization problem over criteria
that measure one or several of these four properties.
In this case, a
grasp is typically defined by the {\em grasp map\/} that transforms
the forces exerted at a set of contact points
to object wrenches~\citep{Murray1994}. The criteria are based on 
geometric, kinematic or dynamic formulations. Analytic formulations
towards grasp synthesis have also been reviewed by~\citet{AB:00}.

Empirical or {\em data-driven\/} approaches rely on sampling 
grasp candidates for an object and ranking them according to a
specific metric. This process is usually based on
some existing grasp experience that can be a heuristic or is generated in
simulation or on a real robot. \citet{Kamon94learningto} refer to this as the 
{\em comparative\/}  and~\citet{Shimoga06} as the {\em knowledge-based\/}
approach. Here, a grasp is
commonly  parameterized by~\cite{mybib:ekvall,Karlsruhe:06}: 
\begin{itemize}
\item the grasping point on the object with which the {\em tool center
  point\/} (TCP) should be aligned,
\item the {\em approach vector\/} which describes
  the 3D angle that the robot hand approaches the grasping point with,
\item the wrist orientation of the robotic hand and
\item an initial finger configuration
\end{itemize} 
Data-driven approaches differ in how the set of grasp candidates is
sampled, how the grasp quality is estimated and how good grasps
are represented for future use. Some methods measure grasp quality
based on analytic formulations, but more commonly they encode
e.g. human demonstrations, perceptual information or semantics. 

\subsection{Brief Overview of Analytic Approaches}
Analytic approaches provide guarantees regarding the criteria that
measure the previously mentioned four grasp properties.
However, these are usually based on
assumptions such as simplified contact models, Coulomb friction and rigid
body modeling~\citep{Murray1994,Handbook}.  Although these
assumptions render grasp analysis practical, inconsistencies and
ambiguities especially regarding the analysis of grasp dynamics are
usually attributed to their approximate nature. 

In this context, \citet{AB:00} identified the problem of finding an
accurate and tractable model of contact compliance as particularly
relevant. This is needed to analyze statically-indeterminate grasps in
which not all internal forces can be controlled. This case arises e.g.
for under-actuated hands or grasp synergies
where the number of the controlled degrees of freedom is fewer than
the number of contact forces. \citet{Prattichizzo:12} model such a
system by introducing a set of springs at the contacts and joints and
show how its dexterity can be analyzed.
\citet{Rosales2012} adopt the same model of compliance to synthesize feasible and prehensile grasps. In this
case, only statically-determinate grasps are considered. The problem
of finding a suitable hand configuration is cast as a constrained
optimization problem in which compliance is introduced to simultaneously 
address the constraints of contact reachability,
object restraint and force controllability. As is the case with many other analytic
approaches towards grasp synthesis, the proposed model is only studied
in simulation where accurate models of the hand kinematics, the object
and their relative alignment are available.

In practice, systematic and random errors are inherent to a robotic
system and are due to noisy sensors and inaccurate models of the robot's 
kinematics and dynamics, sensors or of the object. 
The relative position of object and hand can therefore only be
known approximately which makes an accurate placement of the
fingertips difficult.  In 2000, \citet{AB:00}  identified a lack
of approaches towards synthesizing grasps that are robust to
positioning errors. 
One line of research in this direction explores the concept of {\em independent contact
regions\/} (ICRs) as defined by~\citet{Nguyen88}: a set of regions on
the object in which each finger can be independently placed anywhere
without the grasp loosing the force-closure property. Several examples
for computing them are presented by \citet{Roa2009}
or~\citet{Orebro:10}. Another line of research towards robustness
against inaccurate end-effector positioning makes use of the caging
formulation. \citet{Rodriguez2011} found that there are caging
configurations of a three-fingered manipulator around a planar object
that are specifically suited as a waypoint to grasping it. Once the
manipulator is in such configuration, either opening or closing the
fingers is
guaranteed to result in an equilibrium grasp without the need for
accurate positioning of the fingers. \citet{Seo2012} exploited the
fact that two-fingered immobilizing grasps of an object are always
preceded by a caging configuration. Full body grasps of planar objects
are synthesized by first finding a two-contact caging configuration
and then using additional contacts to restrain the object. Results
have been presented in simulation and demonstrated on a real robot.

Another assumption commonly made in analytic approaches is that
precise geometric and physical models of an object are available to
the robot which is not always the case. In addition, we may not know
the surface properties or friction coefficients,
weight, center of mass and weight distribution. Some of these can be
retrieved through interaction: \citet{Zhang2012} propose to use a
particle filter to simultaneously estimate the physical parameters of
an object and track it while it is being pushed.  The dynamic model of the
object is formulated as a mixed nonlinear complementarity
problem.  The authors show that even when the object is occluded and
the state estimate cannot be updated through visual observation, the
motion of the object is accurately predicted over time.
Although methods like this relax some of the assumptions, they are still limited to 
simulation~\citep{Rodriguez2011,Rosales2012} or consider 2D objects~\citep{Rodriguez2011,Seo2012,Zhang2012}. 

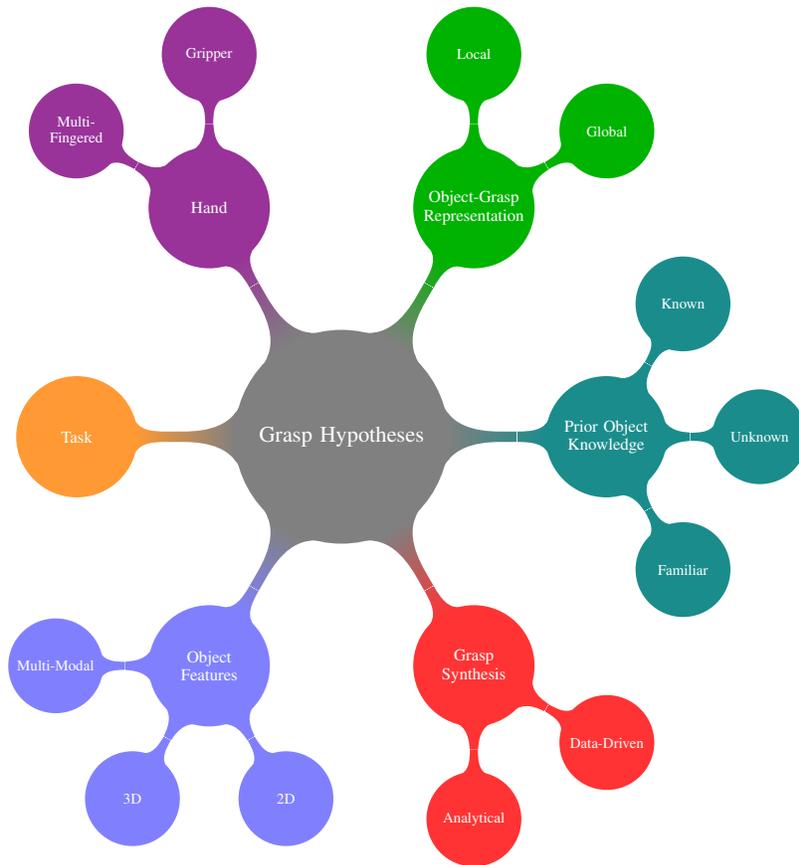
\begin{figure*}[!ht]
\centering
\resizebox{0.6\linewidth}{!}{
\begin{tikzpicture}[mindmap,
  level 3 concept/.append style={sibling angle=60},
  level 4 concept/.append style={sibling angle=60}
  ]  
  \path[mindmap,concept color=black!50!white,text=white] 
  node[concept] {Grasp Hypotheses} 
  [clockwise from=0] 
  child[concept color=teal!90!white] {
    node[concept] {Prior Object Knowledge} 
    [clockwise from=60] 
    child { node[concept] {Known} } 
    child { node[concept] {Unknown} } 
    child { node[concept] {Familiar} }
  } 
  child[concept color=red!80!white] { 
    node[concept] {Grasp Synthesis} 
    [counterclockwise from=-90] 
    child { node[concept] {Analytical} } 
    child { node[concept] {Data-Driven} }
  }
 child[concept color=blue!50!white] {
    node[concept] {Object Features} 
    [clockwise from=-60] 
    child { node[concept] (2d) {2D} } 
    child { node[concept] (3d) {3D} }
    child { node[concept] (multi) {Multi-Modal} }
  } 
  child[concept color=orange!80!white] { node[concept] {Task} }
  child[concept color=violet!80!white] { node[concept] {Hand} 
    [counterclockwise from=90] 
    child { node[concept] {Gripper} } 
    child { node[concept] {Multi-Fingered} }
  }
  child[concept color=green!70!black] { node[concept] {Object-Grasp Representation} 
    [counterclockwise from=30] 
    child { node[concept] {Global} } 
    child { node[concept] {Local} }
  };
\end{tikzpicture}
}

\caption{We identified a number of aspects that influence how the final
  set of grasp hypotheses is generated for an object. The most important
  one is the assumed {\bf prior object knowledge} as discussed in
  Section~\ref{sec:taxonomy}. Numerous different {\bf object-grasp
  representations} are proposed in the literature that are relying on
  {\bf features} of different modalities such as 2D or 3D vision or tactile
  sensors. Either local object parts or the object as a whole are
  linked to specific grasp configurations. {\bf Grasp synthesis} can either
  be analytic or data-driven. The latter is further detailed in
  Fig.~\ref{fig:empericalMinMap}. Very few approaches explicitly
  address the {\bf task} or {\bf hand} kinematics of the robot. }
\label{fig:relatedMindMap}
\end{figure*}

\subsection{Development of Data-Driven Methods}
Up to the year 2000, the field of robotic grasping\footnote{Citation counts for the most influential articles in the field. Extracted from
  \url{scholar.google.com} in October 2013. \citep{Nguyen88}: 733. \citep{AB:00}:
  490. \citep{Canny:92}: 477. \citep{Shimoga06}:
  405. \citep{Kamon94learningto}: 77. \citep{GraspIt}:
  384. \citep{MillerKCA03}: 353. \citep{SVMGrasping}:
  100. \citep{GoldfederALP07}: 110. \citep{Hirzinger:03}:
  95. \citep{Eigen:07}: 96. \citep{diankov_thesis}:
  108. \citep{Morales:04}: 38. \citep{macl08affTRO}: 156.
  \citep{detry2009Human}: 39. \citep{Saxena:06}:
  277. \citep{SaxenaWN08}: 75. \citep{ICVS:08:Shape}:
  40. \citep{LeICRA10}: 21.
  \citep{Bohg2010362}: 43. \citep{Hsiao:10}:
  77. \citep{KlingbeilICRA11}: 26. \citep{Rusu:09:ICRA}: 191.
  \citep{LaiBRF11a}: 58. \citep{TowelFolding:10}:
  75. \citep{pancakes11humanoids}: 39.}  was clearly
dominated by analytic approaches~\citep{Nguyen88,AB:00,Canny:92,Shimoga06}.
Apart from e.g. \citet{Kamon94learningto}, data-driven
grasp synthesis started to become popular with the availability of
GraspIt! \citep{GraspIt} in 2004. Many highly cited
approaches have been developed, analyzed and evaluated in this
or other
simulators~\citep{MillerKCA03,SVMGrasping,GoldfederALP07,Hirzinger:03,Eigen:07,
diankov_thesis}.
These approaches differ in how grasp candidates are sampled from
the infinite space of possibilities. For grasp ranking, they rely 
on classical metrics that are based on analytic
formulations such as the widely used $\epsilon$-metric proposed
in~\citet{Canny:92}. It constructs the {\em grasp
wrench space\/} (GWS) by computing the convex hull over the wrenches
at the contact points between the hand and the object. $\epsilon$
quantifies the quality of a force closure grasp by the radius
of the maximum sphere still fully contained in the GWS.

Developing and evaluating approaches in
simulation is attractive because the environment and its attributes 
can be completely controlled. A large number of experiments can be
efficiently performed without having access to expensive robotics
hardware that would also add a lot of complexity to the evaluation process. 
However, it is not clear if the simulated environment resembles the
real world well enough to transfer methods easily. Only
recently, several articles \citep{skewness:10,Weisz2012,diankov_thesis} 
have analyzed this question and came to the
conclusion that the classic metrics are not good predictors for grasp 
success in the real world. They do not seem to cope well with the challenges arising in
unstructured environments. \citet{diankov_thesis} claims that in
practice grasps synthesized using these metrics tend to be relatively
{\em fragile\/}. \citet{skewness:10} systematically tested a number of
grasps in the real world that were stable according to
classical grasp metrics. Compared to grasps planned by humans and
transferred to a robot by kinesthetic teaching on the same objects,
they under-performed significantly.  A similar study has been conducted
by~\citet{Weisz2012}. It focuses on the ability of the
$\epsilon$-metric to predict grasp stability under object pose
error. The authors found that it performs poorly especially when
grasping large objects.

As pointed out by~\citet{AB:00} and \citet{Handbook}, grasp closure is often
wrongly equated with stability. Closure states the existence
of equilibrium which is a necessary but not sufficient
condition. Stability can only be defined when considering the grasp as
a dynamical system and in the context of its behavior when perturbed
from an equilibrium.  Seen in this light, the results of the above
mentioned studies are not surprising. However, they suggest that there
is a large gap between reality and the models for grasping that are
currently available and tractable.

For this reason, several researchers \citep{Morales:04,macl08affTRO,detry2009Human}
proposed to let the robot learn how to grasp by experience that is gathered 
during grasp execution. Although,
collecting examples is extremely time-consuming, the problem of
transferring the learned model to the real robot is non-existant.
A crucial question is how the object to be grasped is represented and
how the experience is generalized to novel objects. 

\citet{Saxena:06} pushed machine learning approaches for data-driven
 grasp synthesis even further. A simple logistic regressor was trained on 
large amounts of synthetic labeled training data to predict good
grasping points in a monocular image. The authors demonstrated their
method in a household scenario in which a robot emptied a
dishwasher. None of the classical principles
based on analytic formulations were used. This paper spawned a lot of
research \citep{SaxenaWN08,ICVS:08:Shape,LeICRA10,Bohg2010362}
in which essentially one question is addressed: What are the
object features that are sufficiently discriminative to infer a
suitable grasp configuration?

From 2009, there were further developments in the area of 3D
sensing. Projected Texture Stereo was proposed
by~\citet{ProjectedStereo}. This technology is built into the sensor head of the
PR2~\citep{PR2}, a robot that is available to comparatively many
robotics research labs and running on the OpenSource middle-ware ROS~\citep{ROS}.
In 2010, Microsoft released the Kinect~\citep{Kinect}, a highly
accurate depth sensing device based on the technology developed
by~\citet{PrimeSense}. Due to its low price and simple usage, it
became a ubiquitous device within the robotics community. 
Although the importance of 3D data for grasping has been previously
recognized, many new approaches were proposed that
operate on real world 3D data. They are either heuristics that map structures in
this data to grasp configurations 
directly~\citep{Hsiao:10,KlingbeilICRA11} or  they try to detect and
recognize objects and estimate their pose~\citep{Rusu:09:ICRA,lai_icra11a}. 

\subsection{Analytic vs. Data-Driven Approaches}
Contrary to analytic approaches, methods following the data-driven
paradigm place more weight on the object representation and the
perceptual processing, e.g., feature extraction,
similarity metrics, object recognition or classification and pose estimation.
The resulting data is then used to retrieve grasps from some knowledge
base or sample and rank them by comparison to existing grasp experience.
The parameterization of the grasp is less
specific (e.g. an approach vector instead of fingertip positions) and
therefore accommodates for uncertainties in perception and
execution. This provides a natural precursor to reactive
grasping~\citep{Felip2009,Hsiao_09,PastorIROS2011,Hsiao:10,Romano:2011},
which, given a grasp hypothesis, considers the problem of robustly
acquiring it under uncertainty.  
Data-driven methods cannot provide guarantees regarding the
aforementioned criteria of dexterity, equilibrium,
stability and dynamic behaviour~\citep{Shimoga06}. They can only be verified
empirically. However, they form the basis for studying grasp dynamics
and further developing analytic models that better resemble reality.

\subsection{Classification of Data-Driven Approaches}\label{sec:taxonomy}
\citet{Sahbani:2011} divide the data-driven methods based on whether they employ
object features or observation of humans during grasping. 
We believe that this falls short of capturing the diversity of
these approaches especially in terms of the ability to transfer grasp
experience between similar objects and the role of perception in this process.
In this survey, we propose to group data-driven grasp synthesis
approaches based on what they assume to know {\em a priori\/} about
the query object:
\begin{itemize}
\item {\em Known Objects\/}: These approaches assume that the query object
      has been encountered before and that grasps have already been
      generated for it. Commonly, the robot has access to a
      database containing geometric object models that are associated
      with a number of good grasp.  This database is usually built offline and in
      the following will be referred to as an {\em experience
      database\/}. Once the object has been
      recognized, the goal is to estimate its pose and retrieve a
      suitable grasp. 
\item {\em Familiar Objects\/}: Instead of exact identity, the
      approaches in this group assume that the query object is similar to previously
      encountered ones. New objects can be {\em familiar\/} on
      different levels. Low-level similarity can be defined in terms
      of shape, color or texture. High-level similarity
      can be defined based on object category. 
      These approaches assume that new objects
      similar to old ones can be grasped in a similar way. 
      The challenge is to find an object representation and a similarity metric 
      that allows to transfer grasp experience.
\item {\em Unknown Objects\/}: Approaches in this group do not assume
      to have access to object models or any sort of grasp experience.
      They focus on identifying structure or features in sensory data
      for generating and ranking grasp candidates.
      These are usually based on local or global features
      of the object as perceived by the sensor.
\end{itemize}

We find the above classification suitable for surveying the
data-driven approaches since the assumed prior object knowledge
determines the necessary perceptual processing and associated
object representations for generating and ranking grasp
candidates. For known objects, the problems of recognition and pose
estimation have to be addressed. The object is usually represented by a
complete geometric 3D object model. For familiar objects, an object
representation has to be found that is suitable for comparing them to
already encountered object in terms of graspability. For unknown
objects, heuristics have to be developed for directly linking structure in the
sensory data to candidate grasps.
  
  Only a minority of the approaches discussed in this  
  survey cannot be clearly classified to belong to one of these three groups. 
  Most of the included papers use sensor data from the scene to 
  perform data-driven grasp synthesis and are part of a  real robotic
  system that can execute grasps. 

Finally, this classification is well in line with the  research in the
field of neuroscience, specifically, with the theory of the dorsal and
ventral stream in human visual processing \citep{Goodale-92}.  The
{\em dorsal} pathway processes immediate action-relevant features
while the {\em ventral} pathway extracts context- and scene-relevant
information and is related to object recognition. The visual
processing in the ventral and dorsal pathways can be related to the
grouping of grasp synthesis for familiar/known and unknown objects,
respectively. The details of such links are out of the scope of this
paper. Extensive and detailed reviews on the neuroscience of grasping
are offered in \citep{Castiello_2005,Culham_2006,Chinellato_2009}.

\subsection{Aspects Influencing the Generation of Grasp Hypotheses}
\label{genhypos}
\begin{figure}[t]
\centering
\resizebox{0.9\linewidth}{!}{
\begin{tikzpicture}[mindmap,
  level 1 concept/.append style={sibling angle=180},
  ]  
  \path[mindmap, concept color=red!80!white,text=white] 
  node[concept] {Data-Driven Grasp Synthesis} 
  [clockwise from=0] 
  child[concept color=black!30!white, font=\Large] { node[concept] {Heuristics} }
  child[concept color=black!30!white, font=\Large] { node[concept] {Learning}
    [counterclockwise from=120]
    child[font=\large] { node[concept] {Human Demonstration} } 
    child[font=\large] { node[concept] {Labeled Training Data} }
    child[font=\large] { node[concept] {Trial \& Error} }
};
\end{tikzpicture}
}
\caption{Data-driven Grasp Synthesis can either be based on heuristics
  or on learning from data. The data can either be provided in the
  form of offline generated labeled training data, human demonstration
  or through trial and error.}
\label{fig:empericalMinMap}
\end{figure}
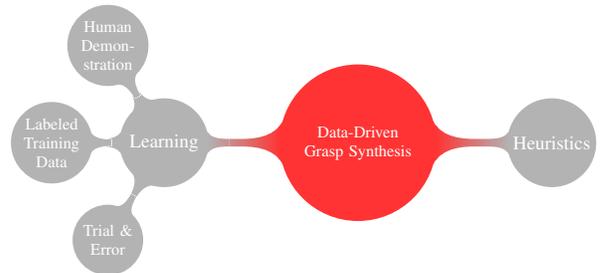

The number of {\em candidate grasps\/} that can be applied to an object is
infinite. To sample some of these candidates and define a quality metric
for selecting a good subset of {\em grasp hypotheses\/}
is the core subject of the approaches reviewed in this survey.
In addition to the prior object knowledge, 
we identified a number of other factors that characterize these
metrics. Thereby, they influence which grasp hypotheses are selected
by a method. Fig.~\ref{fig:relatedMindMap} shows a mind map that structures these
aspects. An important one is how the quality of
a candidate grasp depends on the object, i.e., the {\em object-grasp
representation\/}. Some approaches extract local
object attributes (e.g. curvature, contact area with the hand) around a
candidate grasp. Other approaches take global
characteristics (e.g. center of mass, bounding box) and their relation to a grasp
configuration into account. Dependent on the
sensor device, {\em object features} can be based on 2D or 3D visual data as well
as on other modalities. Furthermore, {\em grasp synthesis\/} can be
analytic or data-driven. We further
categorized the latter in Fig.~\ref{fig:empericalMinMap}: there are
methods for {\em learning\/} either from {\em human demonstrations\/},
{\em labeled examples\/} or {\em trial and error\/}. Other methods rely on various
{\em heuristics\/} to directly link structure in sensory data to candidate grasps.
There is relatively little work on task-dependent grasping. 
Also, the applied robotic hand is usually not in the focus of
the discussed approaches.  We will therefore not examine these two
aspects. However, we will indicate whether an approach takes
the task into account and whether an approach is developed for a
gripper or for the more complex case of a multi-fingered hand.
Table~\ref{tab:known}-\ref{tab:unknown} list all the methods
in this survey. The table columns follow the structure proposed in 
Fig.~\ref{fig:relatedMindMap} and~\ref{fig:empericalMinMap}. 



\section{Grasping Known Objects}
\label{sec:known}
\begin{table}\tiny
\begin{tabularx}{\columnwidth}{p{2cm}|XX|XXX|XXXX|XXXX}
\toprule
 & \multicolumn{2}{p{1cm}}{Object-Grasp Represen.} & \multicolumn{3}{|c}{Object Features} & \multicolumn{4}{|c|}{Grasp Synthesis}& & &\\
\noalign{\smallskip}
\cline{2-14}\noalign{\smallskip}
 & \begin{sideways} Local \end{sideways} & \begin{sideways}
  Global \end{sideways}
& \begin{sideways}2D  \end{sideways}& \begin{sideways}3D \end{sideways}
& \begin{sideways}Multi-Modal\end{sideways}& \begin{sideways}
      Heuristic \end{sideways}& \begin{sideways} Human
      Demo \end{sideways} & \begin{sideways}Labeled
      Data\end{sideways}& \begin{sideways} Trial \&
        Error \end{sideways} & \begin{sideways} Task \end{sideways}
      & \begin{sideways} Multi-Fingered \end{sideways} & \begin{sideways} Deformable \end{sideways} & \begin{sideways} Real Data \end{sideways}\\
\midrule
\citet{Glover:08}              & \V &    & \V &    &    & \V &    &    &    &  &  & \V  & \V \\
\citet{GoldfederALP07}         & \V &    &    & \V &    & \V &    &
&    &  & \V  &   & \\
\citet{MillerKCA03}            & \V &    &    & \V &    & \V &    &
&    &  & \V  &   &\\
\citet{Przybylski_Asfour_2011} & \V &    &    & \V &    & \V &    &
&    &  & \V  &   &\\
\citet{Roa2012}                & \V &    &    & \V &    & \V &    &
&    &  & \V  &   & \V\\
\citet{detry2009Human}         & \V &    &    &    & \V &    & \V &    & \V &  &  &   & \V \\
\citet{detry2010Trial}         & \V &    &    &    & \V & \V &    &    & \V &  &  &   & \V \\
\citet{HuebnerK09}             & \V &    &    &    & \V & \V &    &
&    & \V & \V &   & \V\\
\citet{diankov_thesis}       & \V & \V   &    & \V   &   &  \V  &  &
&    &  & \V  &   &  \\
\citet{skewness:10}            &    & \V &    & \V &    & \V & \V &
&    & \V & \V &   & \V\\
\citet{Hirzinger:03}           &    & \V &    & \V &    & \V &    &
&    &  & \V  &   &  \\
\citet{Brook:2011}             &    & \V &    & \V &    & \V &    &    &    &  &  &   & \V\\
\citet{Eigen:07}               &    & \V &    & \V &    & \V &    &
&    &  & \V  &   & \\
\citet{Romero09}         &    & \V &    & \V   &    &     & \V &    &
&  & \V  &   & \V\\
\citet{Papazov:2012} &    & \V &    & \V  &    & \V    &  & &
 &  & \V &   & \V\\
\citet{Karlsruhe:06}           &    & \V &  &    & \V  &  \V &    &
&    &  & \V  &   & \V\\
\citet{Collet2009}             &    & \V &    &    & \V &  & \V   &
&    &  & \V  &   & \V\\
\citet{Kroemer:2010}           &    & \V &    &    & \V &    & \V &
& \V &  & \V  &   & \V\\
\citet{mybib:ekvall}           &    & \V &    &    & \V &    & \V &
&    &  & \V  &   & \V\\
\citet{TeginAl07} &    & \V &    &   & \V & \V  & \V  & &  &  & \V  &   &
\\
\citet{PastorIROS2011}         &    & \V &    &    &   &     & \V &
&    &  & \V  &   & \V\\
\citet{stulp11learningmotion}  &    & \V &    &   &    &     & \V & &
\V &  & \V  &   & \V\\
\bottomrule
\end{tabularx}
\caption{Data-Driven Approaches for Grasping Known Objects}
\label{tab:known}
\end{table}

If the object to be grasped is known and there is already a database
of grasp hypotheses for it, the problem of finding a feasible grasp
reduces to estimating the object pose and then filtering the
hypotheses by reachability. Table~\ref{tab:known} summarizes all the
approaches discussed in this section.

%

\subsection{Offline Generation of a Grasp Experience Database}
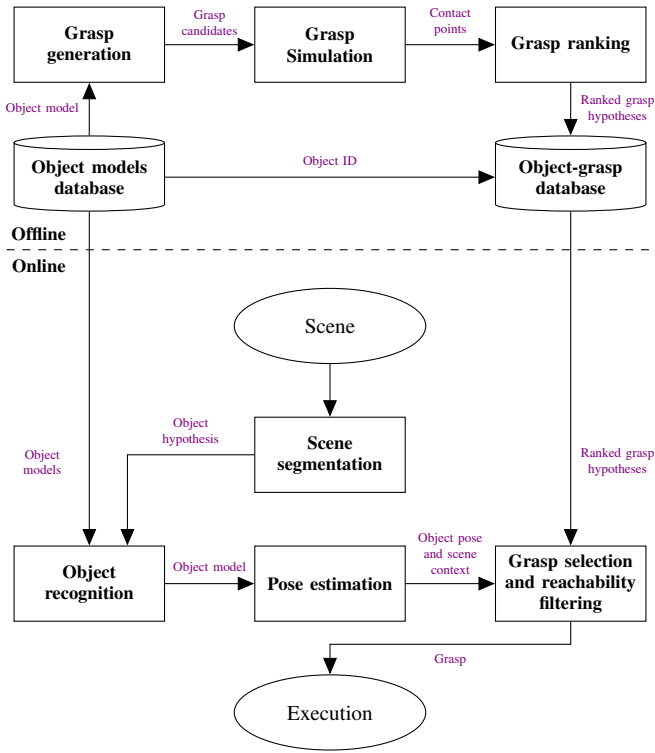
\begin{figure}[!t]
\centering
\begin{tikzpicture} [auto, >=triangle 45]

\matrix[column sep = {3.2cm,between origins}, row sep = 0.7cm, inner sep = 0] (offlinematrix)  
{
  \node[module] (graspgen) {Grasp generation}; & 
  \node[module] (simulation) {Grasp Simulation}; &
  \node[module] (ranking) {Grasp ranking}; \\

  \node[db] (objectdb) {Object models database}; & &
  \node[db] (objectgraspdb) {Object-grasp database}; \\
};

\draw[->] (objectdb) -- node[label] {Object model} (graspgen);
\draw[->] (graspgen) -- node[label] {Grasp candidates} (simulation);
\draw[->] (simulation) -- node[label]  {Contact points} (ranking);
\draw[->] (ranking) -- node[label] (center) {Ranked grasp hypotheses }(objectgraspdb);
\draw[->] (objectdb) -- node[label] {Object ID} (objectgraspdb);

\draw[dashed] ($(offlinematrix.south) + (-4.3cm,-0.5cm)$) --
    node[align = left,font=\scriptsize\bfseries, above, pos = 0.05]  {Offline}
    node[align = left,font=\scriptsize\bfseries, below, pos = 0.05]  {Online}
    ($(offlinematrix.south) + (4.3cm,-0.5cm)$);

\matrix[column sep = {3.2cm,between origins}, row sep = 0.7cm, inner sep = 0, below = 1.0cm of offlinematrix]
{
  & \node [scene] (scene) {Scene}; & \\
  
  & \node [module] (segmentation) {Scene segmentation}; & \\ 
  
  \node [module] (recognition) {Object recognition}; &
  \node [module] (pose) {Pose estimation}; &
  \node [module] (reachability) {Grasp selection and reachability filtering}; \\

  & \node [execution] (execution) {Execution}; & \\
};

\draw [->] (scene) -- (segmentation); 
\draw [->] (segmentation) -| node[label,above, near start] {Object hypothesis} (recognition.45);
\draw [->] (recognition) -- node[label] (lobject) {Object model} (pose);
\draw [->] (pose) -- node[label] {Object pose and scene context} (reachability);
\draw [->] (reachability.south) -- ++(0, -0.3cm) -| node[label, below, near start] {Grasp} (execution.north); 
\draw [->] (objectdb) -- node[label, left, near end] {Object models}(recognition);
\draw [->] (objectgraspdb) -- node[label, right, near end] {Ranked grasp hypotheses} (reachability);

\end{tikzpicture}  
\caption{Typical functional flow-chart for a system with offline
  generation of a grasp database. In the offline phase, every object
  model is processed to generate grasp candidates. Their quality is
  evaluated for ranking. 
  Finally, the list of grasp hypotheses is stored with the
  corresponding object model. In the online phase, the scene is
  segmented to search and recognize object models. If the process
  succeeds, the associated grasp hypotheses are retrieved and
  unreachable ones are discarded. Most of the following approaches can be
  summarized with this
  flowchart. Some of them only implement the offline part. \citep{GoldfederALP07,MillerKCA03,Przybylski_Asfour_2011,Roa2012,HuebnerK09,diankov_thesis,skewness:10,Hirzinger:03,Brook:2011,Eigen:07,Karlsruhe:06,TeginAl07} }

\label{fig:diagramKnownContact}
\end{figure}

First, we look at approaches for generating the experience
database. 
Figs.~\ref{fig:diagramKnownContact} and \ref{fig:diagramKnownHuman}
summarize the typical functional flow-chart of these type of
approaches. 
Each box represents a processing step. Please note, 
that these figures are abstractions that summarize the implementations 
of a number of papers. Most reviewed papers focus on a single module. 
This is also true for similar figures appearing in Sections \ref{sec:unknown} and \ref{sec:familiar}.

\subsubsection{3D Mesh Models and Contact-Level Grasping}\label{sec:known:contactlevel}
Many approaches in this category assume that a 3D
mesh of the object is available. The challenge is then
to automatically generate a set of good grasp hypotheses. This
involves sampling the infinite space of possible hand configurations and
ranking the resulting candidate grasps according to some quality metric. 
The major part of the approaches discussed in the following use force
closure grasps and rank them according to the previously discussed
$\epsilon$-metric. They differ mostly in the way the grasp candidates
are sampled. Fig.~\ref{fig:diagramKnownContact} shows a flow-chart
of which specifically the upper part (Offline) visualizes the data
flow for the following approaches.



\begin{figure}%
  \centering
  \begin{minipage}[!t]{0.453\linewidth}%
    \subfloat[Primitive Shape Decomposition \citep{MillerKCA03}.]{\fbox{\includegraphics[width=\linewidth]{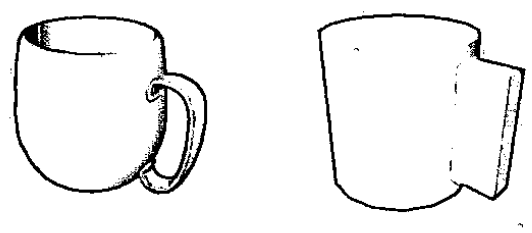}}\label{fig:known:prim}}\\
    \subfloat[Box Decomposition \citep{HuebnerK08}.]{\fbox{\includegraphics[width=\linewidth]{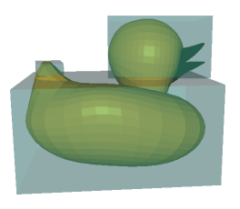}}\label{fig:known:BoxDecomp}}\\
    \subfloat[SQ Decomposition
    \citep{GoldfederALP07}.]{\fbox{\includegraphics[width=\linewidth]{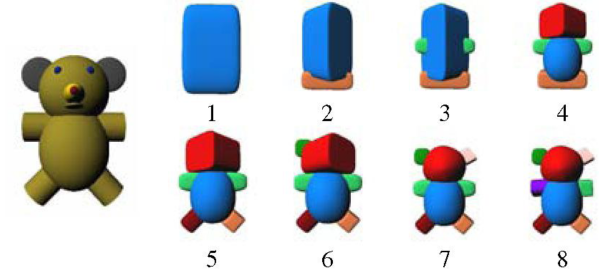}}\label{fig:known:SQs}}\\
    \subfloat[Randomly sampled grasp hypotheses \citep{Hirzinger:03}.]{\fbox{\includegraphics[width=\linewidth]{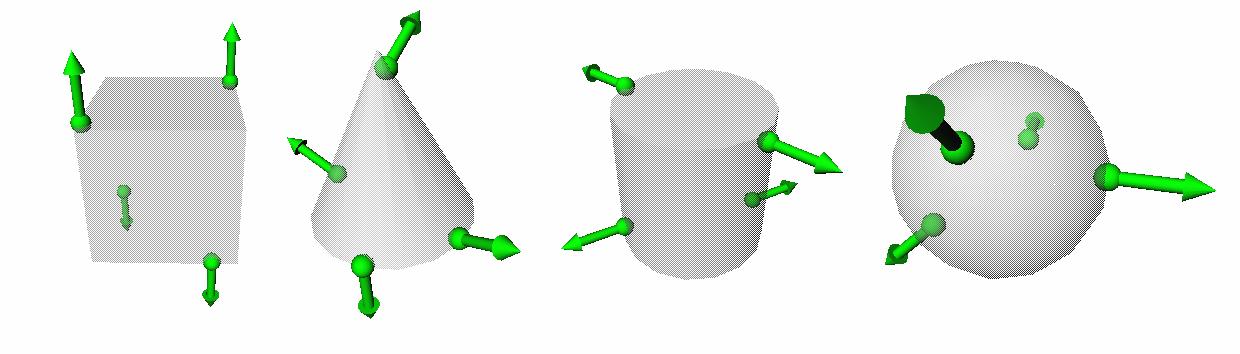}
}\label{fig:known:sampling1}}
  \end{minipage}%
  \qquad
  \begin{minipage}{0.467\linewidth}%
    \subfloat[Green: Centers of a union of spheres. Red: Centers at a
    slice through the
    model \citep{Przybylski_Asfour_2010,Przybylski_Asfour_2011}.]{\fbox{\includegraphics[width=\linewidth]{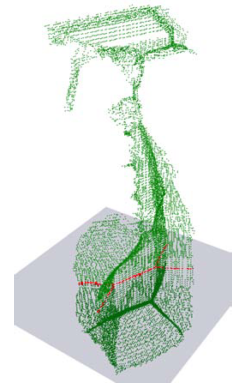}}\label{fig:known:balls}} \\
    \subfloat[Grasp candidate sampled based on surface normals and
    bounding box \citep{Diankov2008}.]{\fbox{\includegraphics[width=\linewidth]{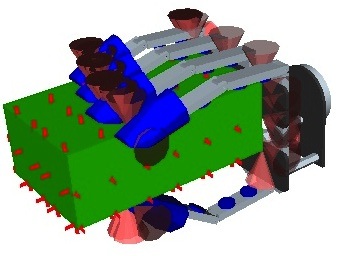}}\label{fig:known:sampling2}}
  \end{minipage}%
  \caption{Generation of grasp candidates through object shape
    approximation with primitives or through
    sampling. \ref{fig:known:prim}) Primitive Shape Decomposition
    \citep{MillerKCA03}. \ref{fig:known:BoxDecomp}) Box Decomposition
    \citep{HuebnerK08}. \ref{fig:known:SQs}) SQ Decomposition
    \citep{GoldfederALP07}. \ref{fig:known:sampling1})
    Randomly sampled grasp hypotheses.\citep{Hirzinger:03}. \ref{fig:known:balls}) Green: Centers of a union of spheres. Red: Centers at a  slice through the
    model \citep{Przybylski_Asfour_2010,Przybylski_Asfour_2011}. \ref{fig:known:sampling2}) Grasp candidate sampled based on surface normals and
    bounding box \citep{Diankov2008}.}%
  \label{fig:known:shapedecomp}
\end{figure}

Some of them approximate the object's shape with a constellation of
primitives such as spheres, cones, cylinders and
boxes as in \citet{MillerKCA03,HuebnerK08} and \citet{Przybylski_Asfour_2011} or superquadrics
(SQ) as in \citet{GoldfederALP07}. These shape primitives are then used to
limit the amount of candidate grasps and thus prune the search tree
for finding the best grasp hypotheses. Examples for
these approaches are shown in
Fig.~\ref{fig:known:prim}-\ref{fig:known:SQs} and Fig.~\ref{fig:known:balls}.
\citet{Hirzinger:03} reduce the number of candidate grasps by
randomly generating a number of them dependent on the object surface
and filter them with a simple heuristic.
The authors show that this approach works well if the goal is not to
find an optimal grasp but instead a fairly good grasp that works well
for `` everyday tasks''. \citet{diankov_thesis} proposes to
sample grasp candidates dependent on the objects bounding box in
conjunction with surface normals. Grasp parameters that are varied are
the distance between the palm of the hand and the grasp point as well
as the wrist orientation.  The authors find that usually a relatively
small amount of 30\% from all grasp samples is in force closure. 
Examples for these sampling approaches are shown in Fig.~\ref{fig:known:sampling1}
and \ref{fig:known:sampling2}.
\citet{Roa2012} present an approach towards synthesizing power grasps
that is not based on evaluating the force-closure property. 
Slices through the object model and perpendicular to
the axes of the bounding box are sampled. The ones that best resemble a
circle are chosen for synthesizing a grasp. 

All these approaches are developed and evaluated in
simulation. As claimed by e.g. \citet{diankov_thesis}, the biggest
criticism towards ranking grasps based on force closure and the
$\epsilon$-metric is that relatively {\em fragile grasps\/} might be
selected. A common approach to filter these, is to add noise to the
grasp parameters and keep only those grasps in which a certain
percentage of the neighboring candidates also yield force closure. 
\citet{Weisz2012} followed a similar approach that focuses in particular on the 
ability of the $\epsilon$-metric to predict grasp stability under object 
pose uncertainty. For a set of object models, the authors used
GraspIt!~\citep{GraspIt} to generate a set of grasp candidates in
force closure. For each object, pose uncertainty is simulated
by perturbing it in three degrees of freedom. Each grasp candidate was
then re-evaluated according to the probability of attaining a force
closure grasp. The authors found that their
proposed metric performs superior especially on large objects. 

\citet{skewness:10} question classical grasp metrics in
principle. The authors systematically tested a number of task-specific grasps in the
real world that were stable 
according to classical grasp metrics. These grasps
under-performed significantly when compared to
grasps planned by humans through kinesthetic teaching on the same
objects and for the same tasks. The authors found that humans 
optimize a {\em skewness\/} metric, i.e., the divergence of alignment
between hand and principal object axes. 

\subsubsection{Learning from Humans}
\begin{figure}[t]
\centering
\begin{tikzpicture}  [auto, >=triangle 45]
\matrix [row sep = 0.7cm , column sep={3.2cm,between origins}] (diamatrix)
{
  & \node[scene] (scene) {Scene: Human Demonstration}; & \\ 
  
  & \node[module] (segmentation) {Scene segmentation}; &
  \node[module] (grasprec) {Grasp recognition/Kinesthetic Teaching};\\

  \node[db] (objectdb) {Object database}; &
  \node[module] (recognition) {Object recognition};   & \\

  & \node[module] (pose) {Pose estimation}; &
  \node[db] (objectgraspdb) {Object-grasp database}; \\ 
};

\draw[->] (scene.south) -- (segmentation);
\draw[->] (segmentation) -- node[label, left] {Object hypothesis} (recognition);
\draw[->] (recognition) -- node[label, left] {Object model} (pose);
\draw[->] (objectdb) -- node[label, below] {Object models} (recognition);

\draw[->] (scene.south) -- ++ (0, -0.3cm) -| (grasprec);
\draw[->] (grasprec) -- node[label, near end] {Grasp hypotheses} (objectgraspdb);
\draw[->] (pose) -- node[label, below] {Object pose \& scene context}(objectgraspdb);

\draw[dashed] ($(diamatrix.south) + (-4.3cm,-0.5cm)$) --
    node[font=\scriptsize\bfseries, above, pos = 0.11]  {Offline learning}
    ($(diamatrix.south) + (4.3cm, -0.5cm)$);

\end{tikzpicture}
\caption{Typical functional flow-chart of a system that learns
  from human demonstration. The robot observes a human grasping 
  a known object. Two perceptual processed are followed in
  parallel. On the left, the object is recognized. On the
  right, the demonstrated grasp configuration is extracted or
  recognized. Finally, object models and grasps are stored together. 
  This process could replace or complement the offline phase
  described in Fig. \ref{fig:diagramKnownContact}. The following
  approaches follow this approach: \citep{detry2009Human,skewness:10,Romero09,Kroemer:2010,PastorIROS2011,stulp11learningmotion}.
}
\label{fig:diagramKnownHuman}
\end{figure}
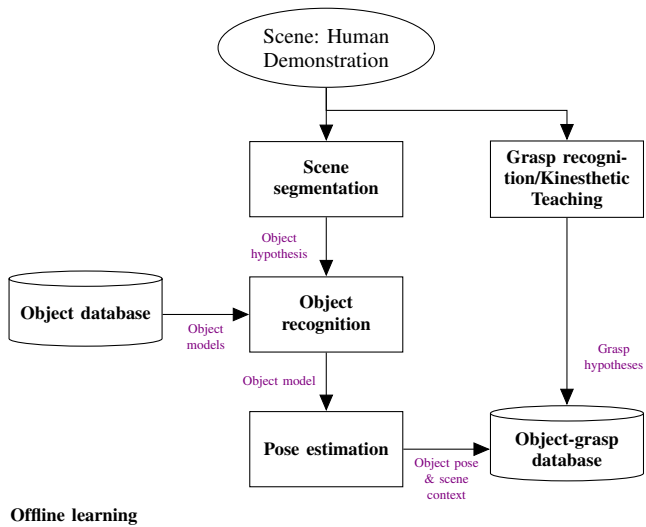

A different way to generate grasp hypotheses is to observe how humans
grasp an object. This is usually done offline following the flow-chart
in Fig.~\ref{fig:diagramKnownHuman}. This process produces an experience database
that is exploited online in a similar fashion as depicted in Fig.~\ref{fig:diagramKnownContact}.

\citet{Eigen:07} exploit results from neuroscience that showed that human hand control takes place in a
space of much lower dimension than the hand's degrees of
freedom. This finding was applied to directly reduce the configuration
space of a robotic hand to find pre-grasp postures. From these so
called {\em eigengrasps\/} the system searches for stable
grasps.

\citet{detry2009Human} model the object as a constellation of local
multi-modal contour descriptors. Four elementary grasping actions are
associated to specific constellations of these features resulting in
an abundance of grasp candidates.  They are
modeled as a non-parametric density function in the space of 6D
gripper poses, referred to as a {\em bootstrap\/} density.
Human grasp examples are used to build an object specific {\em empirical} grasp density from which
grasp hypotheses can be sampled. This is visualized in
Fig.~\ref{fig:known:contourdesc1} and~\ref{fig:known:contourdesc2}.

\citet{Kroemer:2010} represent the object with the same features as
used by~\citet{detry2009Human}. How to grasp specific objects is learned
through a combination of a high-level reinforcement learner and a low
level reactive grasp controller. The learning process is bootstrapped
through imitation learning in which a demonstrated reaching trajectory
is converted into an initial policy. Similar initialization of an object
specific grasping policy is used 
in~\citet{PastorIROS2011} and \citet{stulp11learningmotion}.

\citet{Romero09} present a system for observing humans
visually while they interact with an object. 
A grasp type and pose is recognized and mapped to different robotic hands in a
fixed scheme. 
For validation of the approach in the simulator, 3D object
models are used. This approach has been demonstrated on a humanoid
robot by~\citet{Do2009}. The
object is not explicitly modeled. Instead, it is assumed that human
and robot act on the same object in the same pose.

\begin{figure*}[t]
\centering
\subfloat[Kinesthetic Teaching \citep{Herzog2012}]{\fbox{\includegraphics[height=90pt]{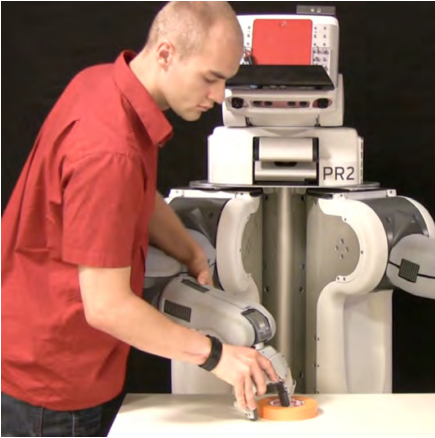}
}\label{fig:known:kines}} \quad
\subfloat[Human-to-robot mapping of grasps using a data
glove\citep{mybib:ekvall}]{\fbox{\includegraphics[height=90pt]{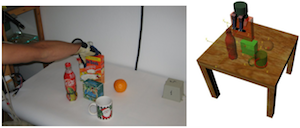}}\label{fig:known:glove}}\quad
\subfloat[Human-to-robot mapping of grasps using visual grasp
recognition \citep{Romero09}]{\fbox{\includegraphics[height=90pt]{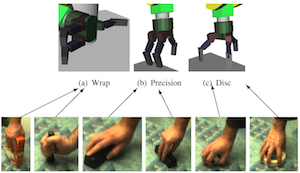}}\label{fig:known:visual}}
\caption{Robot grasp learning from human
  demonstration. \ref{fig:known:kines}) Kinesthetic Teaching
  \citep{Herzog2012}. \ref{fig:known:glove}) Human-to-robot mapping of grasps using a data
glove \citep{mybib:ekvall}. \ref{fig:known:visual}) Human-to-robot mapping of grasps using visual grasp
recognition \citep{Romero09}}
\label{fig:known:humanteaching}
\end{figure*}

In the method presented by \citet{mybib:ekvall}, a human demonstrator
wearing a magnetic tracking device is observed while manipulating a specific
object. The grasp type is recognized and mapped through a fixed schema
to a set of robotic hands. Given the grasp type and the hand, the best
approach vector is selected from an offline trained experience
database. Unlike \citet{detry2009Human} and \citet{Romero09}, the approach vector used by
the demonstrator is not adopted. \citet{mybib:ekvall} assume that the object
pose is known. Experiments are conducted with a simulated
pose error. No physical experiments have been demonstrated.
Examples for the above mentioned ways to teach a robot grasping by demonstration are
shown in
Fig.~\ref{fig:known:humanteaching}. 

\subsubsection{Learning through Trial and Error}
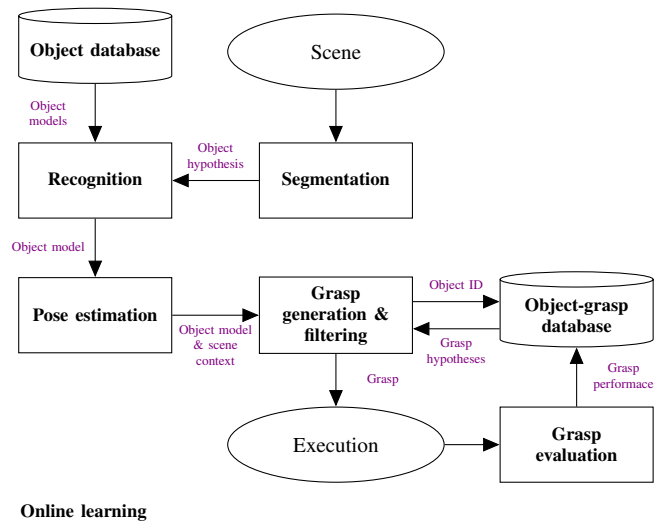
\begin{figure}[!t]
\centering
\begin{tikzpicture}  [auto, >=triangle 45]
\matrix [row sep = 0.7cm , column sep={3.2cm,between origins}] (diamatrix)
{
  \node[db] (objectdb) {Object database}; & 
  \node[scene] (scene) {Scene}; & \\
  
  \node[module] (recognition)  {Recognition}; &
  \node[module] (segmentation) {Segmentation}; & \\

  \node[module] (pose) {Pose estimation}; &
  \node[module] (grasp) {Grasp generation \& filtering}; &
  \node[db] (ogdb){ Object-grasp database}; \\
  
  & \node[execution] (execution) {Execution}; &
  \node[module] (evaluation) {Grasp evaluation}; \\
};

\draw[->] (scene) -- (segmentation);
\draw[->] (segmentation) -- node[label, above] {Object hypothesis} (recognition);
\draw[->] (objectdb) -- node[label, left] {Object models} (recognition);
\draw[->] (recognition) -- node[label, left] {Object model} (pose);
\draw[->] (pose) -- node[label,below] {Object model \& scene context} (grasp);
\draw[->] (grasp) -- node[label] {Grasp} (execution);
\draw[->] (execution) -- (evaluation);
\draw[->] (evaluation) -- node[label, right] {Grasp performace} (ogdb);
\draw[->] (grasp.10) -- node[label,above] {Object ID} (ogdb.170);
\draw[->] (ogdb.190) -- node[label] {Grasp hypotheses} (grasp.350);

\draw[dashed] ($(diamatrix.south) + (-4.3cm,-0.5cm)$) --
    node[font=\scriptsize\bfseries, above, pos = 0.11]  {Online learning}
    ($(diamatrix.south) + (4.3cm, -0.5cm)$);

\end{tikzpicture}
\caption{Typical functional flow-chart of a system that learns through
  trial and error. First, a known object in the scene is segmented and
  recognized. Past experiences with that object are retrieved 
  and a new grasp hypothesis is generated or selected 
  among the already tested ones. After execution of the selected
  grasp, the performance is evaluated and the memory of past
  experiences with the object is updated.The following
  approaches use trial-and-error learning: \citep{detry2009Human,detry2010Trial,Kroemer:2010,stulp11learningmotion}.
}
\label{fig:diagramKnownSelf}
\end{figure}

Instead of adopting a fixed set of grasp candidates for a known
object, the following approaches try to refine them by 
{\em trial and error}. In this case, there is no separation between
offline learning and online exploitation  as can be seen in Fig.~\ref{fig:diagramKnownSelf}.
\citet{Kroemer:2010,stulp11learningmotion} apply
reinforcement learning to improve an initial human demonstration. 
\citet{Kroemer:2010} uses a low-level reactive controller to
perform the grasp that informs the high level controller with reward information.
\citet{stulp11learningmotion} increase the robustness of
their non-reactive grasping strategy by learning shape and goal
parameters of the motion primitives that are used to model a full
grasping action. Through this approach, the robot learns reaching
trajectories and grasps that are robust against object pose uncertainties.
\citet{detry2010Trial} builds a an object-specific empirical grasp density from 
successful grasping trials. This
non-parametric density can then be used to sample grasp hypotheses.

\subsection{Online Object Pose Estimation}
In the previous section, we reviewed different approaches towards
grasping known objects regarding their way to generate and rank
candidate grasps. During online
execution, an object has to be recognized and its pose estimated
before the offline trained grasps can be executed. Furthermore, from
the set of hypotheses not all grasps might be feasible in the
current scene. They have to be filtered by reachability. The lower
part of Fig.~\ref{fig:diagramKnownContact} visualizes the data flow during grasp execution
and how the offline generated data is employed. 

\begin{figure*}[t]
\centering
\subfloat[Object pose estimation of textured and untextured objects in
  monocular images
  \citep{Azad2007}.]{\fbox{\includegraphics[height=70pt]{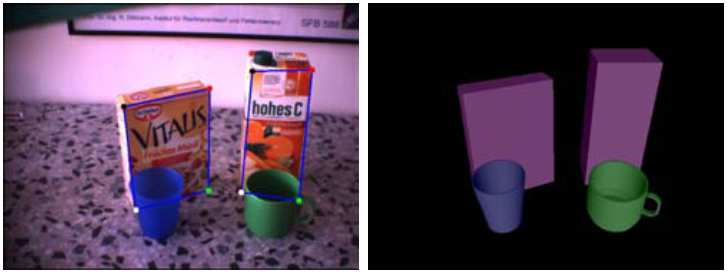}}\label{fig:known:pedram}}\quad
\subfloat[ICP-based object pose estimation from segmented point clouds
\citep{Matei:2010}]{\fbox{\includegraphics[height=70pt]{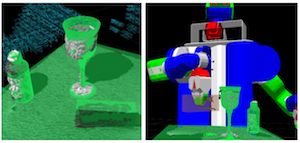}}\label{fig:known:graspPipeline}}\quad
\subfloat[Deformable object detection and pose estimation in monocular
images\citep{Glover:08}]{\fbox{\includegraphics[height=70pt]{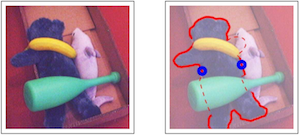}}\label{fig:known:Glover}}\\
\subfloat[Multi-view object representation composed of 2D and 3D features \citep{Collet2009}]{\fbox{\includegraphics[height=81pt]{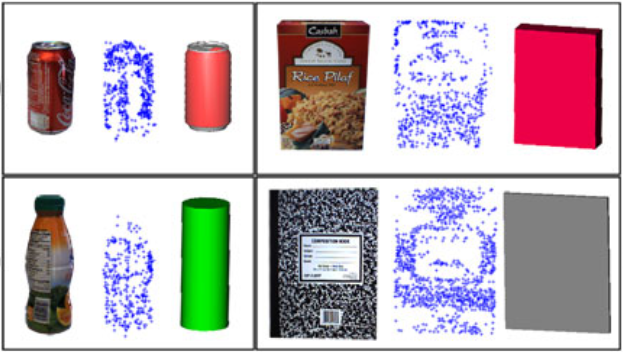}}\label{fig:known:Collet}}\quad
\subfloat[Probabilistic and hierarchical approach towards object pose
  estimation
  \citep{detry2009Pose}]{\fbox{\includegraphics[height=81pt]{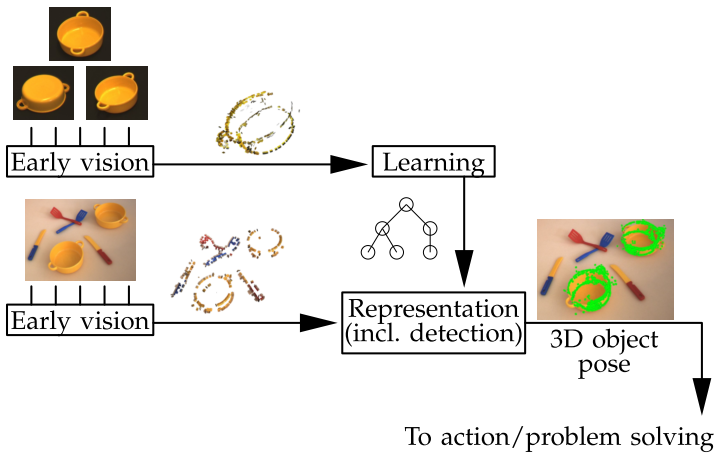}}\label{fig:known:renaud}}\quad
\subfloat[Grasp Candidates linked to groups of local contour descriptors. \citep{detry2009Human}]{\fbox{\includegraphics[height=81pt]{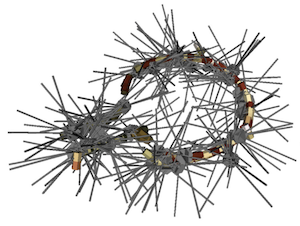}
}\label{fig:known:contourdesc1}}\quad
\subfloat[Empirical grasp density built by trial and error. \citep{detry2009Human}]{\fbox{\includegraphics[height=81pt]{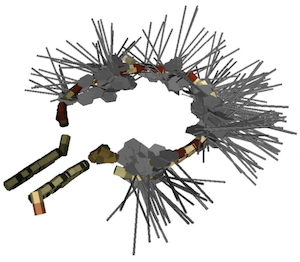}}\label{fig:known:contourdesc2}}
\caption{Object representations for grasping and corresponding methods
  for pose estimation. \ref{fig:known:pedram}) Object pose estimation of textured and untextured objects in
  monocular images \citep{Azad2007}. \ref{fig:known:graspPipeline}) ICP-based object pose estimation from segmented point clouds
\citep{Matei:2010}. \ref{fig:known:Glover}) Deformable object detection and pose estimation in monocular
images\citep{Glover:08}. \ref{fig:known:Collet}) Multi-view object
representation composed of 2D and 3D features
\citep{Collet2009}. \ref{fig:known:renaud}) Probabilistic and hierarchical approach towards object pose
  estimation
  \citep{detry2009Pose}. \ref{fig:known:contourdesc1}) Grasp
  Candidates linked to groups of local contour
  descriptors. \citep{detry2009Human}. \ref{fig:known:contourdesc2}) Empirical grasp density built by trial and error. \citep{detry2009Human}}
\end{figure*}

Several of the aforementioned grasp generation 
methods~\citep{Kroemer:2010,detry2009Human,detry2010Trial} use the
probabilistic approach towards object representation and 
pose estimation proposed by~\citet{detry2009Pose} as visualized in Fig.~\ref{fig:known:renaud}. 
Grasps are either selected by sampling from densities
\citep{detry2009Human,detry2010Trial} or a grasp policy refined from a
human demonstration is applied~\citep{Kroemer:2010}.
\citet{Karlsruhe:06} use the method proposed
by \citet{Azad2007} to recognize an object and estimate its pose from a
monocular image as shown in Fig.~\ref{fig:known:pedram}.
Given this information, an 
appropriate grasp configuration can be selected from a grasp experience 
database that has been acquired offline. The whole system is
demonstrated on the 
robotic platform described in \citet{Asfour2008}.
\citet{HuebnerK09}
demonstrate grasping of known objects on the same humanoid
platform and use the same method for object recognition and pose
estimation. The offline selection of grasp hypotheses is based on a
decomposition into boxes as described in~\citet{HuebnerK08}. Task
constraints are taken into account by reducing the set
of box faces that provide valid approach directions. These constraints
are hard-coded for each task. 
\citet{Matei:2010} propose a robust grasping pipeline in
which known object models are fitted to point cloud
clusters using standard ICP~\citep{ICP:92}. The search space
of potential object poses is reduced by assuming a dominant plane and
rotationally-symmetric objects that are always standing upright as e.g. 
shown in Fig.~\ref{fig:known:graspPipeline}.
\citet{Papazov:2012} demonstrate their previous approach on 3D object recognition and pose estimation
\citep{PapazovB10} in a grasping scenario. Multiple objects in
cluttered scenes can be robustly recognized and their pose
estimated. No assumption is made about the geometry of the scene,
shape of the objects or their pose.


The aforementioned methods assume a-priori known rigid 3D object model.
\citet{Glover:08} consider known deformable objects. Probabilistic
models of their 2D shape are learned offline. The objects can
then be detected in monocular images of cluttered scenes even when
they are partially occluded. The visible object part serve as a basis
for planning a grasp under consideration of the global object
shape. An example for a successful detection is shown in
Fig.~\ref{fig:known:Glover}. 

\citet{Collet_Romea_2011} use a combination of 2D and 3D features as
an object model. Examples for objects from an earlier version of
the system~\citep{Collet2009} are shown in Fig.~\ref{fig:known:Collet}.
The authors estimate the object's pose
in a scene from a single image. The accuracy of their approach is
demonstrated through a number of successful grasps. 

\section{Grasping Familiar Objects}\label{sec:familiar}
The idea of addressing the problem of grasping {\em familiar\/} objects
originates from the observation that many of the objects in the environment
can be grouped together into categories with common characteristics.
In the computer vision community, objects within one category usually
share similar visual properties. These can be, e.g., a common
texture~\citep{Shotton2006} or shape~\citep{Ferrari:08,Belongie:02},
the occurrence of specific local features~\citep{BOW:04,Perona:05} or
their specific spatial
constellation~\citep{LeibeLS06,Lazebnik2006}. These categories are
usually referred to as {\em basic level categories\/} and emerged from
the area of cognitive psychology~\citep{rosch76basic}.

For grasping and manipulation of objects, a more natural
characteristic may be the functionality that they 
afford~\citep{ICVS:08:Shape}: similar objects are grasped in a
similar way or may be used to fulfill the same task (pouring,
rolling, etc). The difficulty is to find a representation that encodes
these common affordances.  Given the
representation, a similarity metric has to be found under which
objects of the same functionality can be considered as alike. The
approaches discussed in this survey
are summarized in Table~\ref{tab:familiar}. All of them employ
learning mechanisms and showed that they can generalize the grasp experience on training
data to new but familiar objects. 

\begin{table}\tiny
\begin{tabularx}{\columnwidth}{p{2cm}|XX|XXX|XXXX|XXXX}
\toprule
 & \multicolumn{2}{p{1cm}}{Object-Grasp Represen.} & \multicolumn{3}{|c}{Object Features} & \multicolumn{4}{|c|}{Grasp Synthesis}& & &\\
\noalign{\smallskip}
\cline{2-14}\noalign{\smallskip}
 & \begin{sideways} Local \end{sideways} & \begin{sideways}
 Global \end{sideways}
 & \begin{sideways}2D  \end{sideways}& \begin{sideways}3D \end{sideways}
 & \begin{sideways}Multi-Modal\end{sideways}& \begin{sideways}
 Heuristic \end{sideways}& \begin{sideways} Human Demo \end{sideways}
 & \begin{sideways}Labeled Data\end{sideways}& \begin{sideways}
 Trial \& Error \end{sideways} & \begin{sideways} Task \end{sideways}
 & \begin{sideways} Multi-Fingered \end{sideways} & \begin{sideways} Deformable \end{sideways} & \begin{sideways} Real Data \end{sideways}\\
\midrule
\citet{SongEHK11} & \V &    &    & \V &    &    &    & \V
&    & \V & \V &   &\\
\citet{Pollard05}         & \V   &    &    & \V &   &    & \V &    &
&  & \V  &   &  \\
\citet{ElKhouryS08}     & \V   &    &   &  \V  &   &    &  &  \V  &
&  & \V  &   &  \\ 
\citet{HuebnerK08}     & \V   &    &   &  \V  &   &    &  &  \V  &
& \V  & \V &   &  \V \\
\citet{Kroemer2012} & \V   &   &   & \V &   &   & \V  &   & \V 
& \V & \V &  & \V \\ 
\citet{detry2012ICRA} & \V   &   &   & \V &   &   &  &  \V &  
&  & \V &  &  \\ 
\citet{detry2013ICRA} & \V   &   &   & \V &   &   &  &  \V &  
&  & \V &  &  \V \\ 
\citet{Herzog2012} & \V   &   &   & \V &   &   & \V &  &  \V
&  & \V &  & \V \\ 
\citet{Ramisa2012} & \V   &   &   & \V &   &   &  & \V & 
&  & \V & \V & \V \\
\citet{BoulariasIROS11}  & \V   &   &    & \V  &   &   &   & \V   &
&  & \V &  & \V \\   
\citet{montesanoRAS2011}  & \V   &   &  \V  &  &   &   &   &   & \V
& & \V &  & \V \\
\citet{ICVS:08:Shape} & \V   &   &  \V &  &   &   & \V  &  \V & 
& \V & &  & \V \\
\citet{Saxena:06} & \V   &   &  \V &  &   &   &   &  \V & 
&  & &  & \V \\ 
\citet{SaxenaWN08}  & \V   &   &   &  & \V  &   &   &  \V & 
&  & \V &  & \V \\ 
\citet{FischingerV12}   & \V   &   &   & \V &   &   &   &  \V & 
&  &  &  & \V \\
\citet{LeICRA10}     & \V   & \V   &   &   & \V  &    &   &  \V  &
&  & \V  &  & \V \\ 
\citet{bergstrom:icvs09} & \V  & \V  &   &   & \V  & \V  &  & \V & 
&  & \V &  & \V \\ 
\citet{HillenbrandR12} & \V & \V  &   & \V  &   &   &  & \V  & 
& \V & \V  &  &  \\ 
\citet{Bohg2010362} &   & \V  &   &   & \V  &   &  & \V & 
&  & \V &  & \V \\ 
\citet{BohgGrasp:2012} &   & \V  &   &   & \V  &   &  & \V & 
& \V & \V &  & \V \\ 
\citet{XiaoIROS08} &  & \V  &   &   & \V  &   &  &  & \V 
&  & \V &  &  \\ 
\citet{Goldfeder:2011} &  & \V  &   &   & \V  &   &  & \V &  
&  & \V &  & \V \\ 
\citet{marton11ijrr} &  & \V  &   &   & \V  & \V  &  & \V  &   
&  & \V &  & \V \\
\citet{RaoICRA10} &  & \V  &   &   & \V  & \V  &  &   & 
&  & \V &  & \V  \\ 
\citet{SpethMS08} &  & \V  &   &   & \V  &   &  &   & \V
&  & \V &  & \V  \\ 
\citet{madry2012ICRA} &  & \V  &   &   & \V  &   &  & \V  & 
& \V & \V &  & \V  \\  
\citet{Kamon94learningto} &  & \V  & \V  &   &   &   &  &   & \V  
&  & &  & \V \\ 
\citet{macl08affTRO} &  & \V  & \V  &   &   &   &  &   & \V  
&  & \V  &  & \V \\ 
\citet{Morales:04} &  & \V  & \V  &   &   &   &  &   & \V  
& & \V  &  & \V \\ 
\citet{SVMGrasping} &  & \V  &   & \V  &   &   &  & \V  & 
& & \V  &  &  \\ 
\citet{DangA12} &  & \V  &   & \V  &   &   &  & \V  & 
& \V & \V  &  & \V \\ 
\bottomrule
\end{tabularx}
\caption{Data-Driven Approaches for Grasping Familiar Objects}
\label{tab:familiar}
\end{table}


\subsection{Discriminative Approaches}
\begin{figure}[!ht]
\centering
\begin{tikzpicture} [auto, >=triangle 45]

\matrix[column sep = {3.2cm,between origins}, row sep = 0.7cm] (offlinematrix)  
{
  \node[db] (samplesdb) {Labeled examples database}; &
  \node[module] (anafeatures) {Feature extraction}; \\
  
  & \node[module] (learning) {Learning features-grasp relation}; \\
};

\draw[->] (samplesdb) -- node[label, above] {Sample} (anafeatures);
\draw[->] (samplesdb) |- node[label, below, near end] {Grasp label of sample} (learning);
\draw[->] (anafeatures) -- node[label, right] {Features} (learning);

\matrix[below = 1cm of offlinematrix, column sep= {3.2cm,between origins}, row sep = 0.7cm] (onlinematrix)
{
  & \node[scene] (scene) {Scene}; & \\
  
  \node[module] (segmentation) {Scene segmentation}; &
  \node[module] (feature) {Feature extraction}; & 
  \node[model] (model) {Learned model features - grasp}; \\
  
  & \node[module] (grasp) {Grasp selection and filtering}; & \\
  
  & \node[execution] (execution) {Execution}; & \\
};

\draw[->, decorate, decoration={coil}] (learning) -- node[label] {Learned model} (model);

\draw[->] (scene.south) -- ++(0, -0.3cm) -| (segmentation);
\draw[->] (segmentation) -- node[label, below] {Segmented cluster} (feature); 
\draw[->] (feature) -- node[label, below] {Features} (model);
\draw[->] (model) |- node[label, near end] {Grasp hypotheses} (grasp);
\draw[->] (segmentation) |- node[label, below, near end] {Scene context} (grasp);
\draw[->] (grasp) -- node[label] {Grasp} (execution);

\draw[dashed] ($(onlinematrix.north) + (-4.3cm,0.5cm)$) --
    node[font=\scriptsize\bfseries, above, pos = 0.105]  {Offline learning}
    node[font=\scriptsize\bfseries, below, pos = 0.05]  {Online}
    ($(onlinematrix.north) + (4.3cm,0.5cm)$);

\end{tikzpicture}
\caption{Typical functional flow-chart of a system that learns from
  labeled examples. In the offline learning phase a database is available
  consisting of a set of objects labeled with grasp configurations and their
  quality. Database entries are analyzed to extract relations between
  specific features and the grasps. The result is a learned model that
  given some features can predict grasp qualities. In the online
  phase, the scene is segmented and features are extracted from the
  scene. Given this, the model outputs a ranked set of promising grasp
  hypotheses.
  Unreachable grasps are filtered out and the best is executed. The
  following approaches use labeled training
  examples: \citep{SongEHK11,Pollard05,ElKhouryS08,HuebnerK08,detry2012ICRA,detry2013ICRA,Ramisa2012,BoulariasIROS11,ICVS:08:Shape,Saxena:06,SaxenaWN08,LeICRA10,bergstrom:icvs09,HillenbrandR12,Bohg2010362,BohgGrasp:2012,XiaoIROS08,Goldfeder:2011,marton11ijrr,madry2012ICRA,SVMGrasping,DangA12}}
\label{fig:diagramFamiliarOffline}
\end{figure}
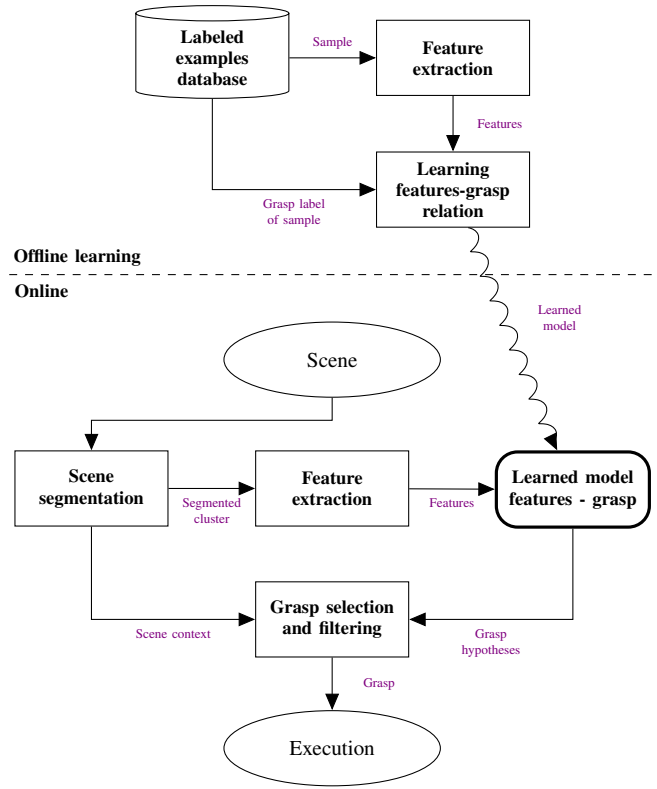

First, there are approaches that learn a discriminative function to 
distinguish between good and bad grasp configurations. 
They mainly differ in what object features are used and thereby in the space over which objects are 
considered similar. Furthermore, they parameterize grasp candidates differently.
Many of them only consider whether a specific part of the object is graspable or not. Others also 
learn multiple contact points or full grasp configurations. A flow-chart for the approaches discussed in the following
is presented in Fig.~\ref{fig:diagramFamiliarOffline}. 

\subsubsection{Based on 3D Data}
\begin{figure}
\centering 
\includegraphics[width=0.8\linewidth]{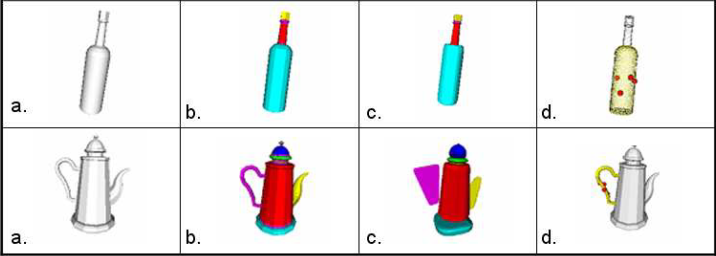}
\caption{a) Object model. b) Part segmentation. c) SQ
  approximation. d) Graspable part and contact points \citep{ElKhouryS08}.}
\label{fig:familiar:Handle}
\end{figure}

\citet{ElKhouryS08} distinguish between graspable and non-graspable 
parts of an object. A point cloud of an object is segmented into parts.
Each part is approximated by a {\em superquadric\/} (SQ). An
artificial neural network (ANN) is used to classify whether or not the
part is prehensile.  The ANN is trained offline on human-labeled
SQs. If one of the object parts is classified as prehensile, an
n-fingered force-closure grasp is synthesized on this object
part. Grasp experience is therefore only used to decide where to
apply a grasp, not how the grasp should be configured. 
These steps are shown for two objects in
Fig.~\ref{fig:familiar:Handle}.

\begin{figure}
\centering 
\includegraphics[width=0.8\linewidth]{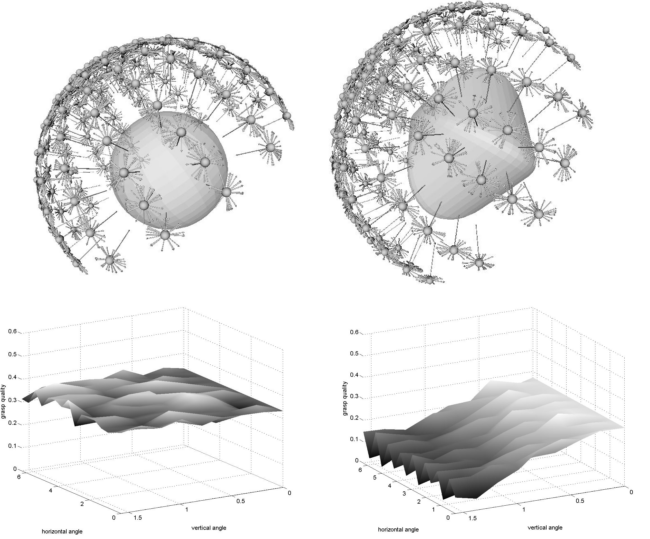}
\caption{Top) Grasp candidates performed on SQ. Bottom) Grasp quality
  for each candidate \citep{SVMGrasping}.}
\label{fig:familiar:SVM}
\end{figure}

\citet{SVMGrasping} approximate an object with a a single
SQ. Given this, their goal is to find a suitable grasp configuration
for a Barrett hand consisting of the approach vector, wrist
orientation and finger spread. A {\em Support Vector Machine\/} (SVM)
is trained on data consisting of feature vectors containing the
SQ parameters and a grasp configuration. They are labeled
with a scalar estimating the grasp quality. This training data is
shown in Fig.~\ref{fig:familiar:SVM}. When feeding the SVM only
with the shape parameters of the SQ, their algorithm searches
efficiently through the grasp configuration space for parameters that
maximize the grasp quality. 

Both of the aforementioned approaches are evaluated in simulation
where the central assumption is that accurate and detailed 3D object models
are available: an assumption not always valid. An SQ
is an attractive 3D representation due to its low number of
parameters and high shape variability. However, it remains unclear whether an
SQ could equally well approximate object shape when given real-world sensory data that is
noisy and incomplete. 

\citet{HuebnerK08} decompose a point cloud into a constellation of boxes. 
The simple geometry of a box
reduces the number of potential grasps significantly. A hand-designed mapping between simple
box features (size, position in constellation) and grasping task is proposed. To decide which
of the sides of the boxes provides a good grasp, an ANN is trained
offline on synthetic data.  The projection of the point cloud inside a
box to its sides provides the input to the ANN.  The training data
consists of a set of these projections from different objects labeled
with the grasp quality metrics. 


\citet{BoulariasIROS11} model an object as a {\em Markov Random Field\/} (MRF) in which 
the nodes are points in a point cloud and edges are spanned between
the six nearest neighbors of a point. The features of a
node describe the local point distribution around that node.  A node in the MRF can carry either
one of two labels: a good or a bad grasp location. The goal
of the approach is to find the maximum a-posteriori labeling of point clouds for new objects.
Very little training data is used which is shown in Fig.~\ref{fig:familiar:MRFGrasp}. A
handle serves as a positive example. The experiments show that this
leads to a robust labeling of 3D object parts that are very similar to
a handle. 
 
Although both approaches~\citep{HuebnerK08,BoulariasIROS11} also rely
on 3D models for learning, the authors show examples for real
sensor data. It remains unclear how well the classifiers
would generalize to a larger set of object categories and
real sensor data.

\citet{FischingerV12} propose a {\em height-accumulated\/} feature that is similar to Haar basis
functions as successfully applied by e.g.~\citet{ViolaJ01} for
face detection. The values of the feature are computed based on the
height of objects above e.g. the table plane.
Positive and negative examples are used to
train an SVM that distinguishes between good and bad grasping points.
The authors demonstrate their approach for cleaning cluttered
scenes. No object segmentation is required for the approach.


\subsubsection{Based on 2D Data}
\begin{figure}[t!]
\centering
\subfloat[One example for each of the eight object classes in training
data in ~\citep {Saxena:06} along with their grasp labels (in
  yellow).]
{
\begin{minipage}{0.6\linewidth}
\vspace{-60pt}
\includegraphics[width=35pt]{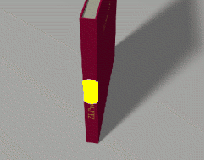}
\includegraphics[width=35pt]{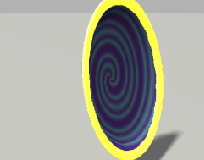}
\includegraphics[width=35pt]{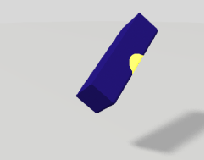}
\includegraphics[width=35pt]{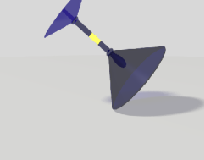}\\
\includegraphics[width=35pt]{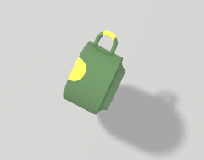}
\includegraphics[width=35pt]{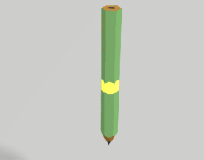}
\includegraphics[width=35pt]{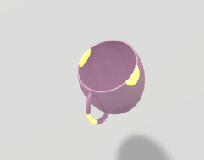}
\includegraphics[width=35pt]{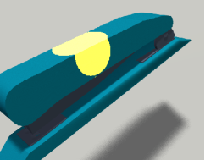}
\end{minipage}
\label{fig:familiar:database}}\quad
\subfloat[Positive (red) and negative examples (blue) for grasping points.~\citep{BoulariasIROS11}]{
\includegraphics[height=65pt]{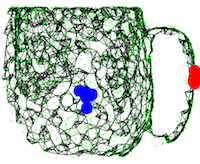}\label{fig:familiar:MRFGrasp}}
\caption{Labeled training data. \ref{fig:familiar:database}) One example for each of the eight object classes in training
data in ~\citep {Saxena:06} along with their grasp labels (in
  yellow). \ref{fig:familiar:MRFGrasp}) Positive (red) and negative
  examples (blue) for grasping points ~\citep{BoulariasIROS11}. }
\end{figure}

There are number of experience-based approaches that avoid the complexity of 3D data and mainly rely 
on 2D data to learn to discriminate between good and bad grasp locations.
\citet{Saxena:06} propose a system that infers a point at where to grasp an 
object directly as a function of its image. The authors apply logistic 
regression to train a grasping point model on labeled synthetic images
of a number of different objects. The classification is based on a
feature vector containing local appearance cues regarding color,
texture and edges of an image patch in several scales and of its
neighboring patches. Samples from the labeled training data
are shown in Fig.~\ref{fig:familiar:database}. The system was used
successfully to pick up objects from a dishwasher after it has been
additionally trained for this scenario.

Instead of assuming the availability of a labeled data set, \citet{montesanoRAS2011} 
allow the robot to autonomously explore which features encode graspability. 
Similar to \citep{Saxena:06}, simple 2D filters are used that can be
rapidly convolved with
an image. Given features from a region, the robot can compute the posterior probability
that a grasp applied to this location will be successful. It
is modeled as a Beta distribution and estimated from the grasping trials executed by the 
robot and their outcome. Furthermore, the variance of the posterior
can be used to guide 
exploration to regions that are predicted to have a high success rate 
but are still uncertain.

Another example of a system involving 2D data and grasp experience is
presented by~\citet{ICVS:08:Shape}. Here, an object is represented by a
composition of prehensile parts.  These so called {\em affordance
cues\/} are obtained by observing the interaction of a person with a
specific object. Grasp hypotheses for new stimuli are inferred by
matching features of that object against a codebook of learned {\em
affordance cues\/} that are stored along with relative object position
and scale. 
How to grasp the detected parts is not solved since hand orientation and finger
configuration are not inferred from the affordance cues.  
Similar to~\citet{BoulariasIROS11}, especially locally very
discriminative structures 
like handles are well detected. 

\subsubsection{Integrating 2D and 3D Data}
Although the above approaches have been demonstrated to work well in specific manipulation
scenarios, inferring a full grasp configuration from 2D data alone is a highly under-constrained problem. 
Regions in the image may have very similar visual features but afford completely different grasps. 
The following approaches integrate multiple complementary modalities, 2D and 3D visual data and their local or 
global characteristics, to learn a function that can take more parameters of a grasp into account.

\citet{SaxenaWN08} extend
their previous work on inferring 2D grasping points by taking the 3d point distribution within a sphere centered
around a grasp candidate into account. This
enhances the prediction of a stable grasp and also allows for the
inference of grasp parameters like approach vector and finger
spread. In earlier work~\citep{Saxena:06}, only downward or
outward grasp with a fixed pinch
grasp configuration were possible.

\begin{figure}[t]
 \centering 
 \includegraphics[width=0.8\linewidth]{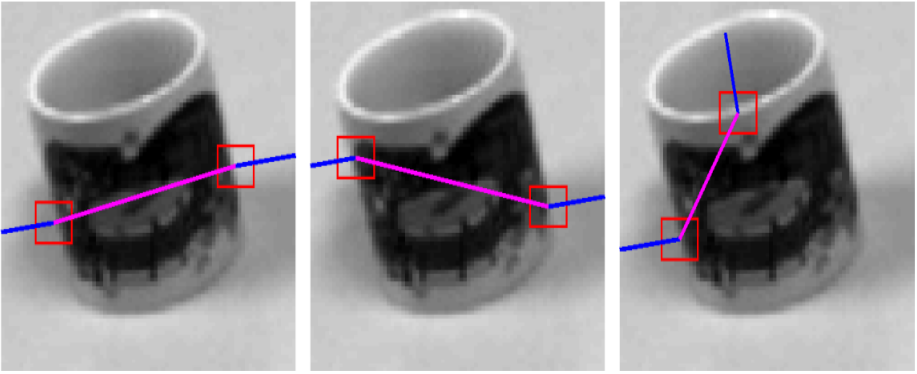}
 \caption{Three grasp candidates for a cup represented by two local
   patches and their major gradient as well as their connecting line. \citep{LeICRA10}}
 \label{fig:familiar:multiContact}
 \end{figure}

\citet{RaoICRA10} distinguish between graspable and
non-graspable object hypotheses in a scene. Using
a combination of 2D and 3D features, an SVM is trained on labeled data
of segmented objects. Among those features are for
example the variance in depth and height as well as variance of the
three channels in the Lab color space. These are
some kind of {\em meta\/} features that are used instead of the
values of e.g. the color channels directly. \citet{RaoICRA10} achieve good
classification rates on object hypotheses formed by segmentation on
color and depth cues. 
\citet{LeICRA10} model grasp hypotheses as
consisting of two contact points. They apply a learning
approach to rank a sampled set of fingertip positions according to
graspability. The feature vector consists of a combination of 2D and
3D cues such as gradient angle or depth variation along the line
connecting the two grasping points. Example grasp candidates are shown
in Fig.~\ref{fig:familiar:multiContact}. 

\begin{figure}
\centering
\parbox{0.35\linewidth}{
\subfloat{\includegraphics[width=\linewidth]{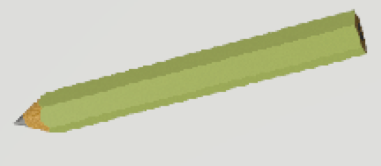}\label{origpencil}}\\
\subfloat{\includegraphics[width=\linewidth]{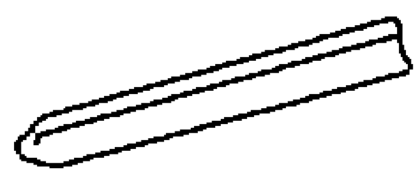}\label{contourpencil}}
}\qquad
\begin{minipage}{0.35\linewidth}
\centering
\subfloat{\includegraphics[width=\linewidth]{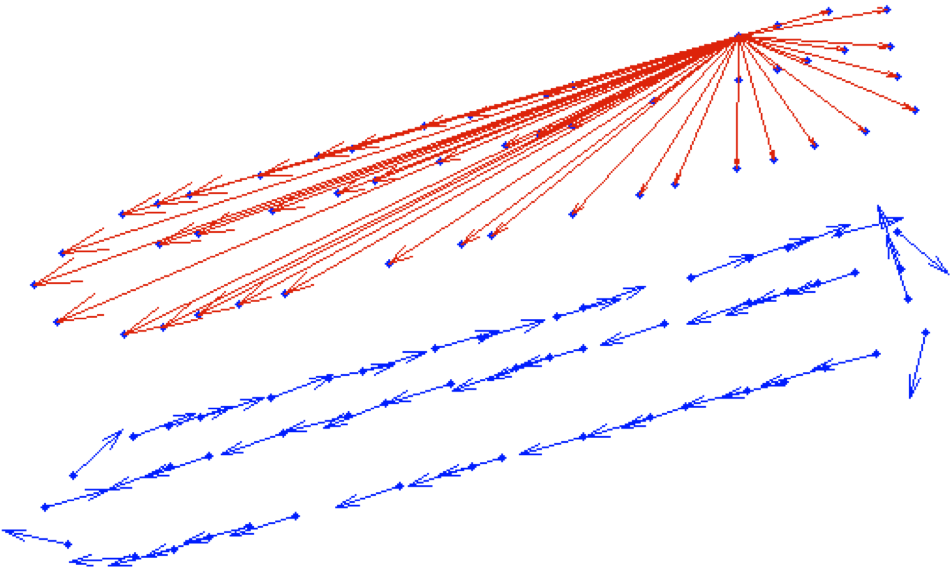}\label{vectors}}\\
\subfloat{\includegraphics[width=0.5\linewidth]{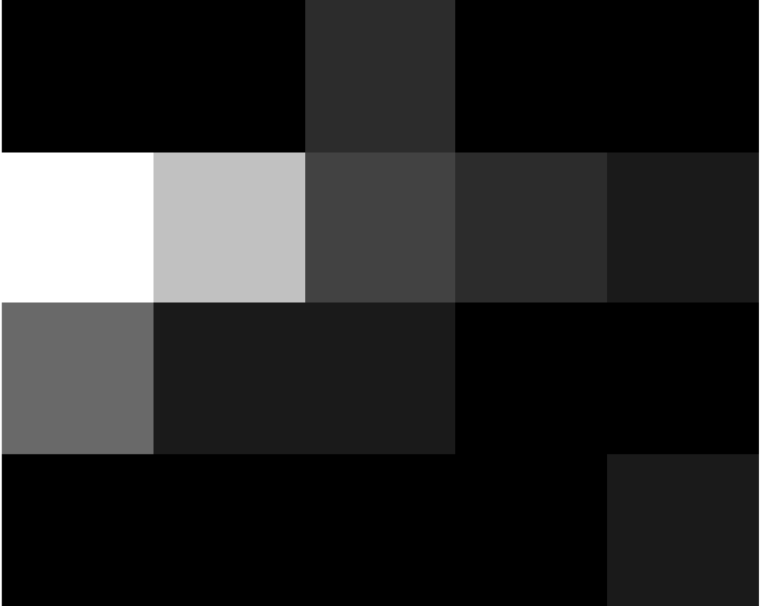}\label{histogram}}
\end{minipage}
\caption{Example shape contexts descriptor for the
  image of a pencil.  \ref{origpencil}) Input
  image. \ref{contourpencil}) Canny edges.  
  \ref{vectors} Top) All vectors from one point to all
  other sample points.  Bottom) Sampled points of
  the contour with gradients. \ref{histogram}) Histogram with four angle and five
  log-radius bins comprising the vectors in \ref{vectors} Bottom)~\citep{Bohg2010362}.}
\label{fig:familiar:SC}
\end{figure}

\citet{Bohg2010362} propose an approach that instead of using local
 features, encodes global 2D object shape. It is represented relative to a potential
grasping point by shape contexts as introduced by~\citet{Belongie:02}.
Fig.~\ref{fig:familiar:SC} shows a potential grasping point and the associated feature. 

\citet{bergstrom:icvs09} see the result of the 2D based grasp
selection as a way to search in a 3D object representation for a full
grasp configuration.  
The authors extend their previous approach \citep{Bohg2010362}
to work on a sparse edge-based object representation.
They show that integrating 3D
and 2D based methods for grasp hypotheses generation
results in a sparser set of grasps with a good quality.

Different from the above approaches, \citet{Ramisa2012} consider
the problem of manipulating deformable objects, specifically folding shirts.
They aim at detecting the shirt collars that exhibit
deformability but also have distinct features.  The authors show that
a combination of local 2D and 3D descriptors works well for this
task.  Results are presented in terms of how reliable
collars can be detected when only a single shirt or several shirts are present in the
scene.

\subsection{Grasp Synthesis by Comparison}
The aforementioned approaches study what kind of features encode
similarity of objects in terms of graspability and learn a
discriminative function in the associated space.  The methods we
review next take an exemplar-based approach in which grasp hypotheses
for a specific object are synthesized by finding the most similar
object or object part in a database to which good grasps are already associated.

\subsubsection{Synthetic Exemplars}\label{sec:comparison:synthetic}
\begin{figure}
\centering 
\includegraphics[width=0.8\linewidth]{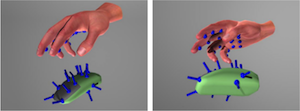}
\caption{Matching contact points between human hand and object \citep{Pollard05}.}
\label{fig:familiar:shapematching}
\end{figure}

\citet{Pollard05} treat the problem of finding a suitable
grasp as a shape matching problem between the human hand and the
object. The approach starts off with a database of human grasp
examples. From this database, a suitable grasp is retrieved when
queried with a new object. Shape features of this object are matched
against the shape of the inside of the available hand postures. An
example is shown in Fig.~\ref{fig:familiar:shapematching}.

\citet{XiaoIROS08} build upon a knowledge base of 3D object types. These are represented 
by Gaussian distributions over very basic shape features, e.g., the
aspect ratio of the object's bounding box, but also over physical
features, e.g. material and weight.  Furthermore, they are annotated
with a set of representative pre-grasps.  To infer a good grasp for a
new object, its features are used to look up the most similar object
type in the knowledge base. If a successful grasp has been synthesized
in this way and it is similar enough to the object type, the mean and
standard deviation of the object features are updated. Otherwise a new
object type is formed in the knowledge base.

While these two aforementioned approaches use low-level shape
features to encode similarity between objects, \citet{DangA12} present
an approach towards {\em semantic\/} grasp planning. In this case, 
{\em semantic\/} refers to both, the object category and the task of a
grasp, e.g. pour water, answer a call or hold and drill.
A {\em semantic affordance map\/} links object features to an approach
vector and to semantic grasp features (task label, joint angles and tactile sensor
readings). For planning a task-specific grasp
on a novel object of the same category, the object features are used to retrieve the optimal approach
direction and associated grasp features. The approach vector serves as a
seed for synthesizing a grasp with the Eigengrasp
planner~\citep{Eigen:07}. The grasp features are used as a reference
to which the synthesized grasp should be similar. 

\citet{HillenbrandR12} frame the problem of transferring functional
grasps between objects of the same category as pose alignment and shape warping. They assume
that there is a source object given on which a
set of functional grasps is defined. Pose clustering is used to align
another object of the same category with it. Subsequently, fingertip contact points
can be transferred from the source to the target
object.  The experimental results are promising. However, they are limited to the category
of cups containing six instances.

All four
approaches~\citep{Pollard05,XiaoIROS08,DangA12,HillenbrandR12} 
compute object features that rely on the availability of 3D object
meshes. The question remains how these ideas could be transferred to the
case where only partial sensory data is available to compute object
features and similarity to already known objects. One idea would be to
estimate full object shape from partial or multiple observations as
proposed by the approaches in Sec.~\ref{sec:unknown:complete} and use
the resulting potentially noisy and uncertain meshes to transfer
grasps. The above methods are also
suitable to create experience databases offline that require only
little labeling. In the case of category-based grasp
transfer \citep{DangA12,HillenbrandR12}  only one object per
category would need to be associated with grasp hypotheses and all the
other objects would only need a category label. No expensive grasp
simulations for many grasp candidates would need to be executed as for the
approaches in Section~\ref{sec:known:contactlevel}. 
\citet{DangA12} followed this idea and demonstrated a few grasp trials on a real
robot assuming that a 3D model of the query object is in the experience database. 

Also \citet{Goldfeder:2011}
built their knowledge base only from synthetic data on which grasps
are generated using the previously discussed Eigengrasp
planner~\citep{Eigen:07}. Different from the above approaches,
observations made with real sensors from new objects are used to look
up the most similar object and its pose
in the knowledge base.  Once
this is found, the associated grasp hypotheses can be executed on the
real object.  Although experiments on a real platform are provided, it
is not entirely clear how many trials have been performed on each
object and how much object pose was varied. 
As discussed earlier, the study conducted by~\citet{skewness:10} suggests that the employed
grasp planner is not the optimal choice for synthesizing
grasps that also work well in the real world. 

\begin{figure}
\centering 
\includegraphics[width=0.9\linewidth]{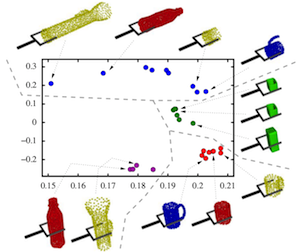}
\caption{ Lower dimensional space in which similar pairs of grasps and
  object parts are close to each other \citep{detry2012ICRA}.}
\label{fig:familiar:Detry}
\end{figure}

\citet{detry2012ICRA} aim at generalizing grasps to 
novel objects by identifying parts to which a grasp has 
already been successfully applied. This look-up is rendered efficient
by creating a lower-dimensional space in which object parts that are
similarly shaped relative to the hand reference frame are close to
each other.  This space is shown in Fig.~\ref{fig:familiar:Detry}.
The authors show that similar grasp to
object part configurations can be clustered in this space and form
prototypical grasp-inducing parts. An extension of this approach is presented
by~\citet{detry2013ICRA} where the authors demonstrate how this
approach can be used to synthesize grasps on novel objects by matching
these prototypical parts to real sensor data.

\subsubsection{Sensor-based Exemplars}
\begin{figure}[!t]
\centering
\begin{tikzpicture} [auto, >=triangle 45]

\matrix[column sep = {3.2cm,between origins}, row sep = 0.7cm] (offlinematrix)  
{  
  & \node[scene] (scene) {Scene} ; & \\
  
  \node[module] (segmentation) {Scene segmentation}; &
  \node[module] (features) {Features extraction}; &
  \node[module] (planner) {Heuristic grasp generation}; \\
  
  & \node[model] (model) {Learned model/ grasp-features database}; & \\
   
  & \node[module] (reachability) {Reachability filtering}; &  \\
  
  & \node[execution] (execution) {Execution}; &
  \node[module] (evaluation) {Grasp evaluation}; \\
};

\draw[->] (scene.south) --++ (0, -0.3cm) -| (segmentation);
\draw[->] (segmentation) -- node[label] {Segmented cluster} (features);
\draw[->] (features) -- node[label] {Features} (planner);
\draw[->] (planner.south) --++ (0, -0.3cm) -| node[label, below, near start] {Grasp candidates} (model.50); 
\draw[->] (features) -- node[label, left] {Features} (model);
\draw[->] (model) -- node[label, left] {Grasp Hypotheses} (reachability);
\draw[->] (segmentation) |- node[label, near end, below] {Scence context} (reachability);
\draw[->] (reachability) -- node[label, left] {Grasp} (execution);
\draw[->] (execution) -- (evaluation);
\draw[->] (evaluation) |- node[label, below, near end] {Model update} (model);

\end{tikzpicture}
\caption{Typical functional flow-chart of a system that learns from
  trial and error. No prior knowledge about objects is assumed. 
  The scene is segmented to obtain object clusters and relevant
  features are extracted.  A heuristic module produces grasp
  candidates from these features. These candidates are 
  ranked using a previously learned model or based on comparison to 
  previous examples. The resulting grasp hypotheses are filtered 
  and one of them is finally executed. The performance of the
  execution is evaluated and the 
  model or memory is updated with this new experience. The following
  approaches can be summarized by this flow chart: \citep{Kroemer2012,Herzog2012,montesanoRAS2011,SpethMS08,Kamon94learningto,macl08affTRO,Morales:04}}
\label{fig:diagramFamiliarSelf}
\end{figure}
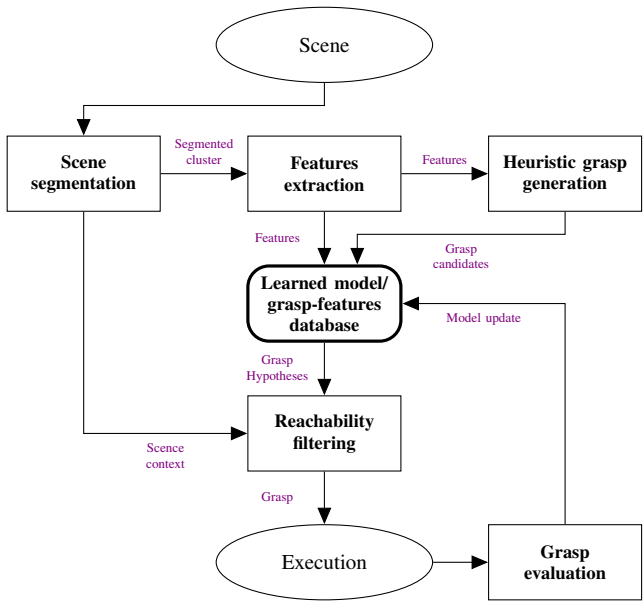

The above mentioned approaches present
promising ideas towards generalizing prior grasp experience to
new objects. However, they are using 3D object models to construct
the experience database. In this section, we review methods that generate a
knowledge base by linking object representations from real sensor data
to grasps that were executed on a robotic platform. Fig.~\ref{fig:diagramFamiliarSelf}
visualizes the flow of data that these approaches follow.

\citet{Kamon94learningto} propose one of the first
approaches towards generalizing grasp experience to novel
objects. Their aim is to learn a function $f: Q \rightarrow G$ that
maps object- and grasp-candidate-dependent quality parameters $Q$ to a
grade $G$ of the grasp.  An object is represented by its 2D
silhouette, its center of mass and main axis. The grasp is represented
by two parameters $f_1$ and $f_2$ from which in combination with the
object features the fingertip positions can be computed. 
Learning is bootstrapped by the offline generation
of a knowledge database containing grasp parameters along with their
grade. This knowledge database is then updated while the robot
gathers experience by grasping new objects. The system is restricted
to planar grasps and visual processing of top-down views on
objects. It is therefore questionable how robust this approach is to
more cluttered environments and strong pose variations of the
object. 

\begin{figure}
\centering
\subfloat{\fbox{\includegraphics[width=0.3\linewidth]{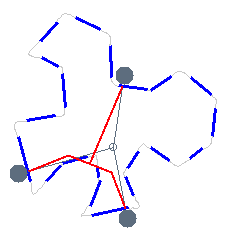}}\label{morales1}} \qquad
\subfloat{\fbox{\includegraphics[width=0.3\linewidth]{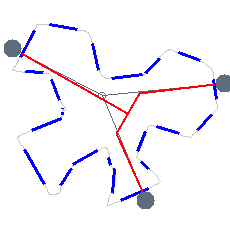}}\label{morales2}}
\caption{ \ref{morales1}) Successful grasp configuration for this object. \ref{morales2})
  Unsuccessful grasp configuration for the same object \citep{Morales:04}.}
\label{fig:unknown:morales}
\end{figure}

\citet{Morales:04} use visual feedback to infer
successful grasp configurations for a three-fingered hand.  The authors take the hand kinematics into
account when selecting a number of planar grasp hypotheses
directly from 2D object contours. To predict which of these grasps is
the most stable one, a {\em k-nearest neighbour\/} (KNN) approach is applied in connection with a
grasp experience database. The experience database is built during a 
trial-and-error phase executed in the real world. Grasp hypotheses
are ranked dependent on their outcome. Fig.~\ref{fig:unknown:morales}
shows a successful and unsuccessful grasp configuration for one object. The approach is restricted to
planar objects. \citet{SpethMS08} showed that their earlier 2D
based approach~\citep{Morales:04} is also applicable when considering
3D objects. The camera is used to explore the object and retrieve
crucial information like height, 3D position and pose. However, all
this additional information is not applied in the inference and final
selection of a suitable grasp configuration. 

\begin{figure}
\centering 
\includegraphics[width=0.8\linewidth]{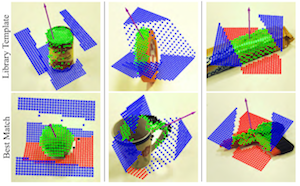}
\caption{Example query and matching templates \citep{Herzog2012}.}
\label{fig:familiar:Alex}
\end{figure}

The approaches presented by~\citet{Herzog2012} and \citet{Kroemer2012}
also maintain a 
database of grasp examples. They combine learning by trial and error on real world 
data with a part based representation of the object. There is no
restriction of object shape. Each of them bootstrap the 
learning by providing the robot with a set of positive example
grasps. However, their part representation and matching is very different. 
\citet{Herzog2012} store a set of local templates of the parts of the
object that have been in contact with the object during the human
demonstration. Given a segmented object point cloud, its 3D convex hull is constructed.
A template is a height map that is aligned with one polygon of this
hull.  Together with a grasp hypotheses, they serve as
positive examples. If a local part of an object is similar to a
template in the database, the associated grasp hypothesis is
executed. Fig.~\ref{fig:familiar:Alex} shows example query templates and the matched
template from the database.
In case of failure, the object part is added as a negative
example to the old template. In this way, the similarity metric can
weight in similarity to positive examples as well as dissimilarity to
negative examples. The proposed approach is evaluated on a large set of different 
objects and with different robots.

\citet{Kroemer2012} use a pouring task to demonstrate the
generalization capabilities of the proposed approach to similar objects. 
An object part is represented as a set of points weighted according to
an isotropic 3D Gaussian with a given standard deviation.
Its mean is manually set to define a part that is relevant to the
specific action.
When shown a new object, the goal of the approach
is to find the sub-part that is most likely to afford the demonstrated
action. This probability is computed by kernel logistic regression whose result 
depends on the weighted similarity between the considered sub-part
and the example sub-parts in the database. The weight vector is learned given the 
current set of examples. This set can be extended with new parts after action execution.
\citet{Herzog2012} and \citet{Kroemer2012} both do not adapt the similarity metric itself under which a new object 
part is compared to previously encountered examples. Instead the probability of success is estimated taken 
all the examples from the continuously growing knowledge base into account.

\subsection{Generative Models for Grasp Synthesis}
Very little work has been done on learning generative models of the
whole grasp process. These kind of approaches identify common
structures from a number of examples instead of finding a decision
boundary in some feature space or directly comparing to previous
examples under some similarity metric. \citet{macl08affTRO} provide one
example in which affordances are encoded in terms of an action that is
executed on an object and produces a specific effect. The problem of
learning a joint distribution over a set of variables is posed as
structure learning in a Bayesian network framework. Nodes in this
network are formed by object, action and effect features that the
robot can observe during execution. Given 300 trials, the robot learns
the structure of the Bayesian network. Its validity is demonstrated in
an imitation game where the robot observes a human executing one of
the known actions on an object and is asked to reproduce the same
observed effect when given a new object. Effectively, the robot has to
perform inference in the learned network to determine the action with
the highest probability to succeed.

\begin{figure}
\centering 
\includegraphics[width=0.7\linewidth]{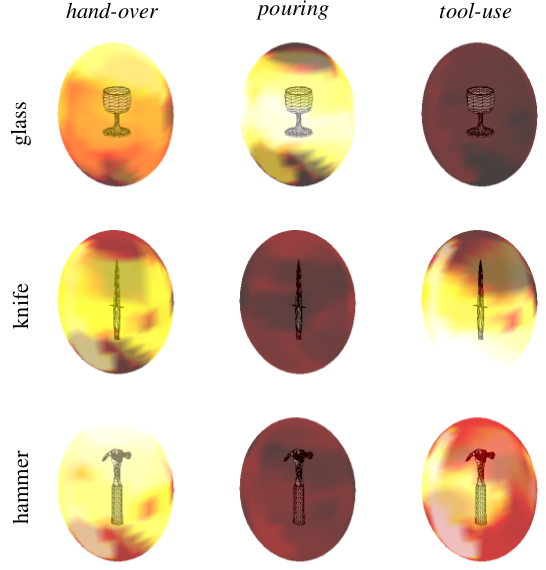}
\caption{Ranking of approach vectors for different objects given a
  specific task. The brighter an area the higher the rank. The darker
  an area, the lower the rank \citep{SongEHK11}.}
\label{fig:familiar:Dan}
\end{figure}

\citet{SongEHK11} approach the problem of inferring a full grasp configuration for an object
given a specific task.  As in~\citep{macl08affTRO}, the joint
distribution over the set of variables influencing this choice is
modeled as a Bayesian network. Additional variables like task, object
category and task constraints are introduced. The structure of this
model is learned given a large number of grasp examples generated in
GraspIt! and annotated with grasp quality metrics as well as
suitability for a specific task. The authors exploit non-linear
dimensionality reduction techniques to find a discrete representation
of continuous variables for efficient and more accurate structure
learning. The effectiveness of the method is demonstrated on the
synthetic data for different inference tasks. The learned quality of
grasps on specific objects given a task is visualized in
Fig.~\ref{fig:familiar:Dan}. 

\subsection{Category-based Grasp Synthesis}
Most of the previously discussed approaches link low-level information
of the object to a grasp.  Given that a novel object is similar in
shape or appearance to a previously encountered one, then it is
assumed that they can also be grasped in a similar way. However,
objects might be similar on a different level.  Objects in a household
environment that share the same functional category might have a
vastly different shape or appearance. However they still can be
grasped in the same way. In Section~\ref{sec:comparison:synthetic}, we have already mentioned
the work by~\citet{DangA12,HillenbrandR12} in which task-specific
grasps are synthesized for objects of the same category. The authors
assume that the category is known a-priori. In the following, we review methods that
generalize grasps to familiar objects by first determining their
category.

\citet{Composite09IROS} use different 3D sensors and a thermo camera
for performing object categorization. Features of the segmented point cloud and
the segmented image region are extracted to train a Bayesian Logic
Network for classifying object hypotheses as either boxes, plates,
bottles, glasses, mugs or silverware.  A modified approach is presented
in \citep{marton11ijrr}.  A layered 3D object descriptor is used for
categorization and an approach based on the 
{\em Scale-invariant feature transform\/} (SIFT)~\citep{Lowe:1999} is applied for view based
object recognition. To increase robustness of the categorization, the
examination methods are run iteratively on the object hypotheses. A
list of potential matching objects are kept and reused for
verification in the next iteration. Objects for which no matching
model can be found in the database are labeled as novel. Given that an
object has been recognized, associated grasp hypotheses can be
reused. These have been generated using the technique presented
in~\citep{Marton10IROS}.

\citet{SongEHK11} treat object
category as one variable in the Bayesian
network. \citet{madry2012ICRA} demonstrate how the category of an
object can be robustly detected given multi-modal visual descriptors
of an object hypothesis. This information is fed into the Bayesian
Network together with the desired task. A full hand configuration
can then be inferred that obeys the task
constraints. \citet{BohgGrasp:2012} demonstrate this approach on the
humanoid robot ARMAR III~\citep{Asfour2008}. For robust object
categorization the approach by \citet{madry2012ICRA} is integrated
with the 3D-based categorization system by \citet{WalterICRA11}. The
pose of the categorized object is estimated with the approach
presented by~\citet{Aitor:2011}. Given this, the inferred grasp
configuration can be checked for reachability and executed by the
robot.

Recently, we have seen an increasing amount of new approaches towards
pure 3D descriptors of objects for categorization. Although, the
following methods look promising, it has not been shown yet that they
provide a suitable base for generalizing grasps over an object
category. \citet{Rusu:10:IROS,Rusu09ICCV-WS} provide extensions
of \citep{Rusu:09:ICRA} for either recognizing or categorizing objects
and estimating their pose relative to the viewpoint. While
in~\citep{Rusu09ICCV-WS} quantitative results on real data are
presented,~\citep{Rusu:10:IROS} uses simulated object point clouds
only. \citet{LaiBRF11a} perform object category and instance
recognition. The authors learn an instance distance using the database
presented in~\citep{lai_icra11a}. A combination of 3D and 2D features
is used. 
\citet{Gonzalez-AguirreHRABD11} present a shape-based object categorization system. 
A point cloud of an object is reconstructed by fusing partial views.
Different descriptors (capturing global and local object shape) in combination with standard 
machine learning techniques are studied. Their performance is evaluated on real data.

\section{Grasping Unknown Objects}
\label{sec:unknown}
If a robot has to grasp a previously unseen object, we refer to it
as {\em unknown\/}. Approaches towards grasping known objects are
obviously not applicable since they rely on the assumption that
an object model is available. The approaches in this
group also do not assume to have access to other kinds of grasp
experiences. Instead, they propose and analyze heuristics that directly
link structure in the sensory data to candidate grasps.

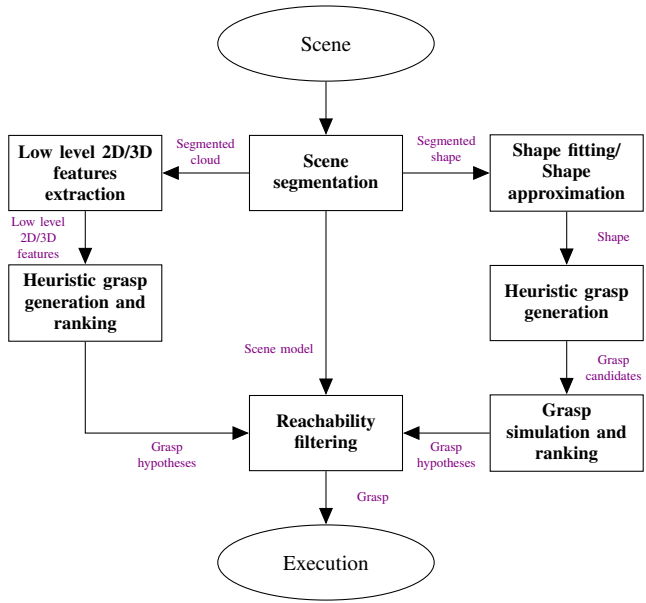
\begin{figure}[!t]
\centering
\begin{tikzpicture} [auto, >=triangle 45]

\matrix [column sep = {3.2cm,between origins}, row sep = 0.7cm ] 
{
  & \node[scene] (scene) {Scene}; & \\

  \node [module] (features) {Low level {2D/3D} features extraction}; &
  \node [module] (segmentation) {Scene segmentation}; & 
  \node [module] (shape) {Shape fitting/ Shape approximation}; \\ 
  
  \node [module] (synthesisA) {Heuristic grasp generation and ranking}; & &
  \node [module] (synthesisB) {Heuristic grasp generation}; \\

  & \node [module ] (reachability) {Reachability filtering}; &
  \node [module] (ranking) {Grasp simulation and ranking}; \\
  
  & \node [execution]   (execution) {Execution}; & \\
};

\draw [->] (scene) -- (segmentation);
\draw [->] (segmentation) -- node[label, above]{Segmented shape} (shape);
\draw [->] (shape)-- node[label,right]{Shape} (synthesisB); 
\draw [->] (synthesisB) -- node[label, right] {Grasp candidates}(ranking); 
\draw [->] (ranking) -- node[label] {Grasp hypotheses} (reachability);
            
\draw [->] (segmentation) -- node[label,above] {Segmented cloud}(features);
\draw [->] (features) -> node[label, left] {Low level {2D/3D} features} (synthesisA); 
\draw [->] (synthesisA)|- node[label, , below, near end]{Grasp hypotheses} (reachability); 
            
\draw [->] (segmentation) -- node [label, left, text width = 1cm, near end] {Scene model} (reachability);
\draw [->] (reachability) --  node [label, right] {Grasp} (execution);

\end{tikzpicture}

\caption{Typical functional flow-chart of a grasping system for
  unknown objects. The scene is perceived and segmented to obtain 
  object hypotheses and relevant perceptual features. Then the system
  follows either the right or left pathway. On the left, low level
  features are used to generate heuristically a set of grasp
  hypotheses.  On the right, a mesh model approximating the 
  global object shape is generated from the perceived features.
  Grasp candidates are then sampled and executed in a
  simulator. Classical analytic grasp metric are used to rank the
  grasp candidates. Finally non reachable grasp hypotheses are
  filtered out, and the best ranked grasp hypothesis is executed. 
  The following approaches use the left pathway: \citep{ObjectBirth09,
  Popovic:11,Richtsfeld08,TowelFolding:10,Hsiao:10,Brook:2011,Holz:11,KlingbeilICRA11,Klank:10,Morales:05}. The
  following approaches estimate a full object model: \citep{Bone:08,bohg:icra11,Marton10IROS,Lippiello:2013,dunes08,Kehoe2012}}
\label{fig:diagramUnknown}
\end{figure}

There are various ways to deal with sparse, incomplete and
noisy data from real sensors such as stereo cameras: we divided the approaches into methods that i)~approximate
the full shape of an object, ii)~methods that generate grasps based on
low-level features and a set of heuristics, and iii)~methods that rely
mostly on the global shape of the partially observed object
hypothesis.  The reviewed approaches are summarized in
Table~\ref{tab:unknown}. A flow chart that visualizes the data flow in
the following approaches is shown in Fig.~\ref{fig:diagramUnknown}. 

\begin{table}\tiny
\begin{tabularx}{\columnwidth}{p{2cm}|XX|XXX|XXXX|XXXX}
\toprule
 & \multicolumn{2}{p{1cm}}{Object-Grasp Represen.}
 & \multicolumn{3}{|c}{Object Features} & \multicolumn{4}{|c|}{Grasp
 Synthesis}& & &\\
\noalign{\smallskip}
\cline{2-14}\noalign{\smallskip}
 & \begin{sideways} Local \end{sideways} & \begin{sideways}
 Global \end{sideways}
 & \begin{sideways}2D  \end{sideways}& \begin{sideways}3D \end{sideways}
 & \begin{sideways}Multi-Modal\end{sideways}& \begin{sideways}
 Heuristic \end{sideways}& \begin{sideways} Human Demo \end{sideways}
 & \begin{sideways}Labeled Data\end{sideways}& \begin{sideways}
 Trial \& Error \end{sideways} & \begin{sideways} Task \end{sideways}
 & \begin{sideways} Multi-Fingered \end{sideways} & \begin{sideways} Deformable \end{sideways} & \begin{sideways} Real Data \end{sideways}\\
\midrule
\citet{ObjectBirth09}   & \V &    &    &      & \V &  \V  &   &    &
 &  &  &   & \V \\
\citet{Popovic:11}      & \V &    &    &      & \V &  \V  &   &    &
& & \V  &   & \V \\
\citet{Bone:08}         & \V &    &    &  \V  &    &  \V  &   &    &
&  &  &   & \V \\
\citet{Richtsfeld08}    & \V &    &    &  \V  &    &  \V  &   &    &
&  &  &   & \V \\
\citet{TowelFolding:10}  & \V   &   &    & \V  &   &  \V  &   &    &
&  & &  \V & \V \\
\citet{Hsiao:10}        & \V & \V &    &  \V  &    &  \V  &   &    &
&  &  &   & \V \\
\citet{Brook:2011}      & \V & \V &    &  \V  &    &  \V  &   &    &
&   & &   & \V \\
\citet{bohg:icra11}     &    & \V &    &  \V  &    &  \V  &   &    &
&   & \V &   & \V \\
\citet{Holz:11}         &    & \V &    &  \V  &    &  \V  &   &    &
&   & &   & \V \\
\citet{KlingbeilICRA11} &    & \V &    &  \V  &    &  \V  &   &    &
&   & &   & \V \\
\citet{Klank:10}        &    & \V &    &  \V  &    &  \V  &   &    &
&   & \V &   & \V \\
\citet{Marton10IROS}    &    & \V &    &  \V  &    &  \V  &   &    &
&   & \V &   & \V \\
\citet{Lippiello:2013}     &    & \V &  & \V     &    &  \V  &   &    &
&   & \V &   & \V \\
\citet{dunes08}         &    & \V & \V &      &    &  \V  &   &    &
&   & &   & \V \\
\citet{Kehoe2012}       &    & \V & \V &      &    &  \V  &   &    &
&   &  &   & \V \\
\citet{Morales:05}      &    & \V & \V &      &    &  \V  &   &    &
&   & \V &   & \V \\
\bottomrule
\end{tabularx}
\caption{Data-Driven Approaches for Grasping Unknown Objects}
\label{tab:unknown}
\end{table}


\subsection{Approximating Unknown Object Shape}\label{sec:unknown:complete}
\begin{figure}
\centering
\subfloat{\includegraphics[height=60pt]{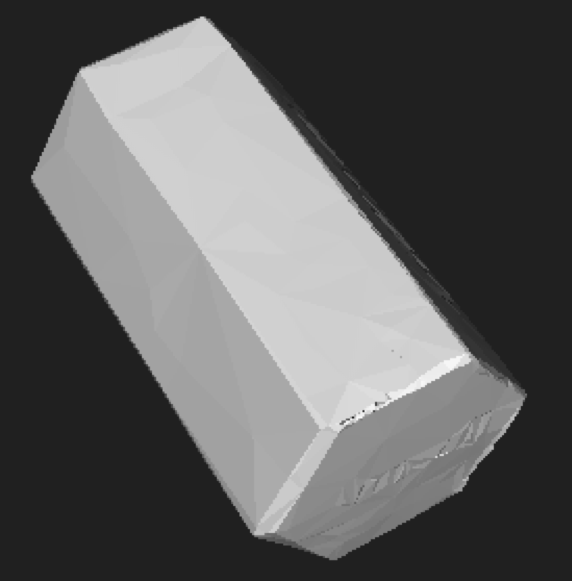}\label{realmesh}}
\subfloat{\includegraphics[height=60pt]{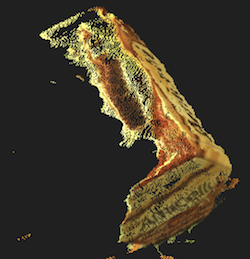}\label{origcloud}}
\subfloat{\includegraphics[height=60pt]{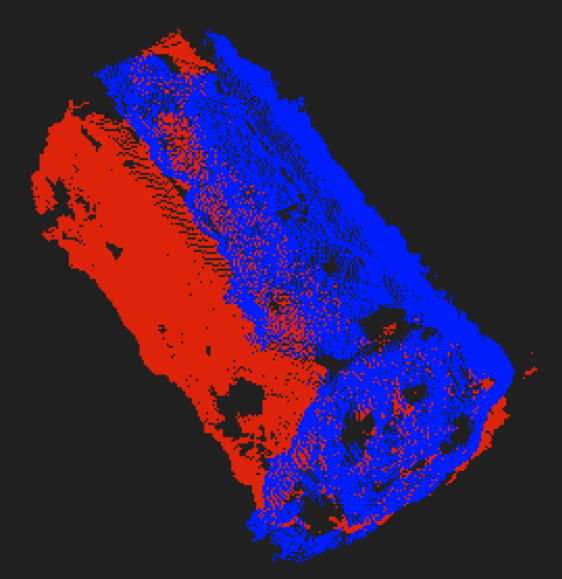}\label{cloudhypo}}
\subfloat{\includegraphics[height=60pt]{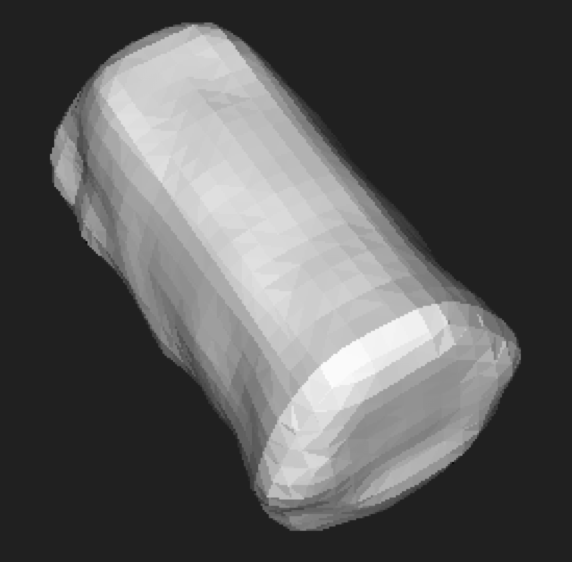}\label{meshmir}}
\caption{ Estimated full object shape by assuming symmetry. \ref{realmesh}) Ground Truth Mesh. \ref{origcloud}) Original Point
  Cloud. \ref{cloudhypo}) Mirrored Cloud with Original Points in Blue and
  Additional Points in Red.\ref{meshmir}) Reconstructed Mesh \citep{bohg:icra11}.}
\label{fig:unknown:symmetry}
\end{figure}

One approach towards generating grasp hypotheses for unknown objects
is to approximate objects with shape primitives.  \citet{dunes08}
approximate an object with a quadric whose minor axis is
used to infer the wrist orientation. The object centroid serves as the
approach target and the rough object size helps to determine the hand
pre-shape.  The quadric is estimated from multi-view measurements of
the global object shape in monocular images.
\citet{Marton10IROS} show how grasp selection
can be performed exploiting symmetry by fitting a curve to a cross
section of the point cloud of an object.  For grasp planning, the
reconstructed object is imported to a simulator. Grasp candidates are
generated through randomization of grasp parameters on which then the
force-closure criteria is evaluated.
\citet{RaoICRA10} sample grasp points from the surface of a segmented
object. The normal of the local surface at this point serves as a
search direction for a second contact point. This is chosen to be at
the intersection between the extended normal and the opposite side of
the object. By assuming symmetry, this second contact point is assumed
to have a contact normal in the direction opposite to the normal of
the first contact point.  
\citet{bohg:icra11} propose a related
approach that reconstructs full object shape assuming planar 
symmetry which subsumes all other kinds of symmetries.  It takes the
complete point cloud into account and not only a local patch. Two simple methods
to generate grasp candidate on the resulting
completed object models are proposed and evaluated. An example for an
object whose full object shape is approximated with this approach is 
shown in Fig.~\ref{fig:unknown:symmetry}.

\begin{figure}
\centering 
\subfloat{\includegraphics[height=46pt]{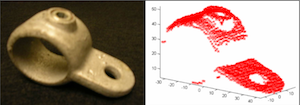}\label{shapecarving1}}
\subfloat{\includegraphics[height=46pt]{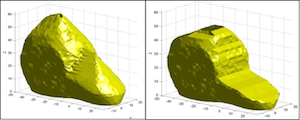}\label{shapecarving2}}
\caption{Unknown object shape estimated by shape carving. \ref{shapecarving1} Left) Object
  Image. Right) Point cloud. \ref{shapecarving2} Left ) Model from
  silhouettes.  Right) Model merged with point cloud data \citep{Bone:08}.}
\label{fig:unknown:shapeCarving}
\end{figure}

Opposed to the above mentioned techniques, \citet{Bone:08} make
no prior assumption about the shape of the object. They apply
shape carving for the purpose of grasping with a parallel-jaw
gripper. After obtaining a model of the object, they search for a pair
of reasonably flat and parallel surfaces that are best suited for this
kind of manipulator. An object reconstructed with this method is 
shown in Fig.~\ref{fig:unknown:shapeCarving}. 

\citet{Lippiello:2013} present a related approach for grasping an
unknown object with a multi-fingered hand. The authors first record a number of
views from around the object. Based on the object bounding box in each view, a polyhedron is
defined that overestimates the visual object hull and is then approximated by a
quadric. A pre-grasp shape is defined in which
the fingertip contacts on the quadric are aligned with its two
minor axes. This grasp is then refined
given the local surface shape close to the contact point.
This process is alternating with the refinement of the object shape through an
elastic surface model. The quality of
the grasps is evaluated by classic metrics. As previously discussed,
it is not clear how well these metrics predict the outcome of a
grasp.

\subsection{From Low-Level Features to Grasp Hypotheses}
\begin{figure*}[t]
\centering 
\subfloat[Generation of grasp candidates from local surface features
  and evaluation in
  simulation. \citep{Popovic:11}]{\fbox{\includegraphics[height=120pt]{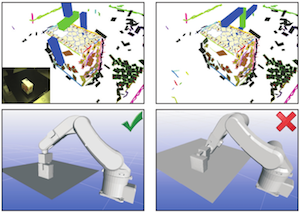}}\label{fig:unknown:surflets}}\quad
\subfloat[Generated grasp hypotheses on point cloud clusters and
  execution
  results.\citep{Hsiao:10}]{\fbox{\includegraphics[height=120pt]{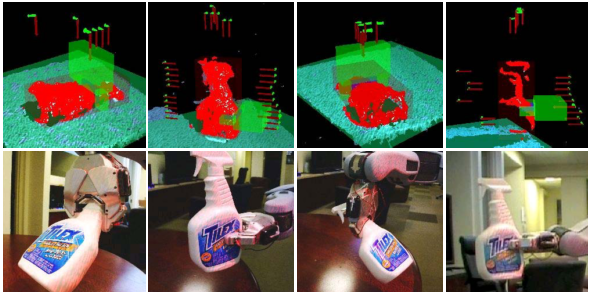}}\label{fig:unknown:contact}}\quad
\subfloat[Top) Grasping a towel. Bottom) Re-grasping a
towel.
\citep{TowelFolding:10}]{\begin{minipage}[!t]{0.153\linewidth}%
\vspace{-115pt}
\fbox{\includegraphics[width=\linewidth]{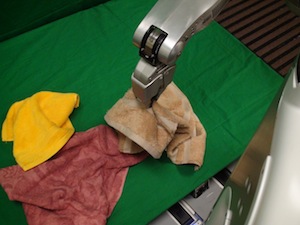}}\\
\fbox{\includegraphics[width=\linewidth]{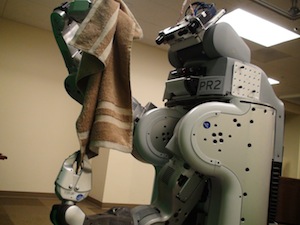}}
\end{minipage}
\label{fig:unknown:towel}}
\caption{Generating and ranking grasp hypotheses from local object
  features. \ref{fig:unknown:surflets}) Generation of grasp candidates from local surface features
  and evaluation in
  simulation \citep{Popovic:11}. \ref{fig:unknown:contact}) Generated grasp hypotheses on point cloud clusters and
  execution
  results \citep{Hsiao:10}. \ref{fig:unknown:towel}) Top) Grasping a
  towel from the table. Bottom) Re-grasping a
towel for unfolding \citep{TowelFolding:10}.}
\label{fig:unknown:summary}
\end{figure*}

A common approach is to map low-level 2D or 3D visual features to a predefined set of 
grasp postures and then rank them dependent on a set criteria.
\citet{ObjectBirth09} use a stereo camera to
extract a representation of the scene. Instead of a raw point cloud,
they process it further to obtain a sparser model consisting of local
multi-modal contour descriptors.
Four elementary grasping actions are
associated to specific constellations of these features. With the help
of heuristics, the large number of resulting grasp hypotheses is
reduced. \citet{Popovic:11} present an extension of this system that
uses local surfaces and their interrelations to propose and
filter two and three-fingered grasp hypotheses. The feasibility of the
approach is evaluated in a mixed real-world and simulated environment.
The object representation and the evaluation in simulation is
visualized in Fig.~\ref{fig:unknown:surflets}.

\citet{Hsiao:10} employ several heuristics for generating
grasp hypotheses dependent on the shape of the segmented point
cloud. These can be grasps from the top, from the side or applied to
high points of the objects. The generated hypotheses are then ranked
using a weighted list of features such as for example number of points
within the gripper or distance between the fingertip and the center of
the segment. Some
examples for grasp hypotheses generated in this way are shown in
Fig.~\ref{fig:unknown:contact}.

\begin{figure}
\centering 
\includegraphics[width=0.8\linewidth]{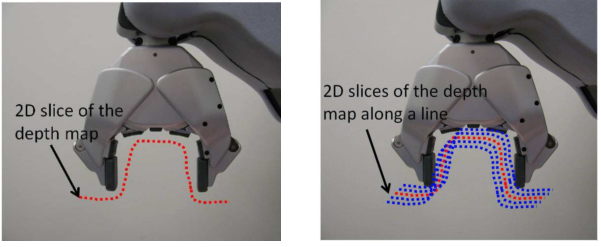}\qquad
\caption{PR2 gripper and associated grasp pattern \citep{KlingbeilICRA11}.}
\label{fig:unknown:Klingbeil}
\end{figure}

The main idea presented by \citet{KlingbeilICRA11} is to search for a
pattern in the scene that is similar to the 2D cross section of the
robotic gripper interior. This is visualized in
Fig.~\ref{fig:unknown:Klingbeil}. The idea is similar to the work
by~\citet{Pollard05} as shown in
Fig.~\ref{fig:familiar:shapematching}. However, in this work the
authors do not rely on the availability of a complete 3D object model.
A depth image serves as the input to the
method and is sampled to find a set of grasp hypotheses. These are
ranked according to an objective function that takes pairs of these
grasp hypotheses and their local structure into account. 

\citet{TowelFolding:10} propose a method for grasping and folding
  towels that can vary in size and are arranged in unpredictable configurations.
Different from the approaches discussed above, the objects are
deformable. The authors propose a border detection 
methods that relies on depth discontinuities and then fit corners to
border points. These then serve as grasping points. Examples for grasping a towel
are shown in Fig.~\ref{fig:unknown:towel}. Although this approach is
applicable to a family of deformable objects, it does not
detect grasping points by comparing to previously encountered grasping
points. Instead it directly links local structure to a grasp. For this
reason, we consider it as an approach towards grasping unknown objects.

\subsection{From Global Shape to Grasp Hypothesis}
\begin{figure}[t]
\centering 
\subfloat[Simplified hand model and grasp parameters to be optimized. \citep{Klank:10}]{\fbox{\includegraphics[height=68pt]{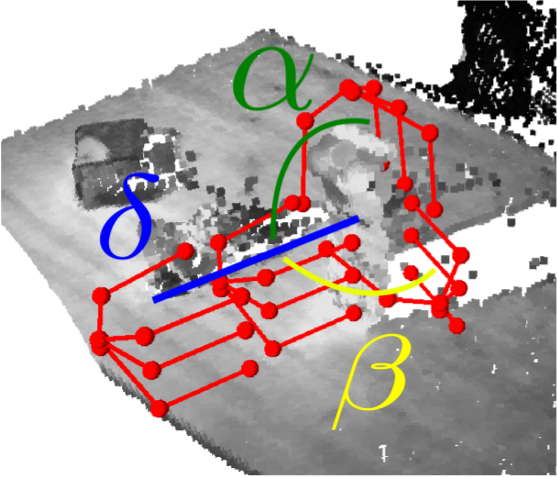}}\label{fig:unknown:optimize}}\quad
\subfloat[Planar object shape uncertainty model Left) Vertices and center of mass with Gaussian
  position uncertainty ($\sigma = 1$). Right) 100 samples of perturbed
  object models. \citep{Kehoe2012}]{\fbox{\includegraphics[height=68pt]{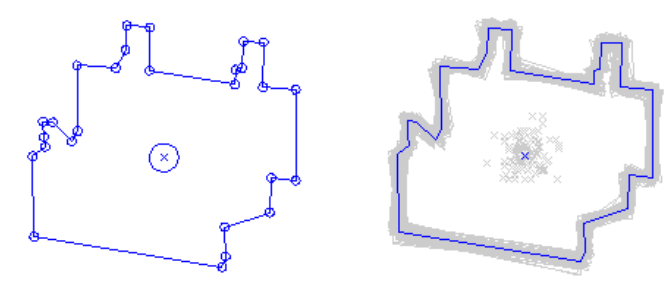}}\label{fig:unknown:push}}
\caption{Mapping global object shape to
  grasps. \ref{fig:unknown:optimize}) Simplified hand model and grasp
  parameters to be
  optimized \citep{Klank:10}. \ref{fig:unknown:push}) Planar object shape uncertainty model Left) Vertices and center of mass with Gaussian
  position uncertainty ($\sigma = 1$). Right) 100 samples of perturbed
  object models \citep{Kehoe2012}.
}
\label{fig:unknown:global}
\end{figure}

Other approaches use the global shape of an object to infer one good
grasp hypothesis. \citet{Morales:05} extracted the {2D} 
silhouette of an unknown object from an image and computed two 
and three-fingered grasps taking into account the kinematics constraints of the hand. 
\citet{Richtsfeld08} use a
segmented point cloud from a stereo camera. They search for a suitable grasp with a simple
gripper based on the shift of the top plane of an object into its
center of mass. A set of heuristics is used for selecting promising
fingertip positions.
\citet{Klank:10}  model the object as a 3D Gaussian. For choosing a
grasp configuration, it optimizes a criterion in which the distance
between palm and object is minimized while the distance between
fingertips and the object is maximized. The simplified model of the
hand and optimization variables
are shown in Fig.~\ref{fig:unknown:optimize}.


\citet{Holz:11} generate grasp hypotheses based on eigenvectors of
the object's {\em footprints\/} on the table. Footprints refer
to the 3D object point cloud projected onto the supporting
surface.

\citet{Kehoe2012} assume  an overhead view of the object and
approximate its shape with an extruded polygon. The goal is to
synthesize a zero-slip push grasp with a parallel jaw gripper given
uncertainty about the precise object shape and the position of its
center of mass. For this purpose, perturbations of the initial shape
and position of the centroid are sampled. For an example of this, see
Fig.~\ref{fig:unknown:push}. For each of these samples,
the same grasp candidate is evaluated. Its quality depends on how
often it resulted in force closure under the assumed model of object shape
uncertainty.

\section{Hybrid Approaches}\label{sec:hybrid}

\begin{figure}
\centering 
\includegraphics[width=\linewidth]{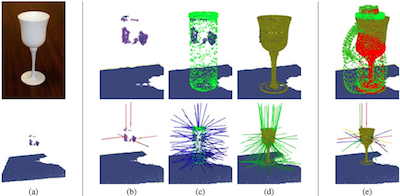}
\caption{a) Object and point cloud. b,c,d) Object representation and
grasp hypotheses. e) Overlaid representations and list of consistent
grasp hypotheses \citep{Brook:2011,Kaijen:ICRAWS:2011}.}
\label{fig:familiar:Brook}
\end{figure}

There are a few data-driven grasp synthesis methods that cannot
clearly be classified as using only one kind of prior knowledge.  One
of these approaches has been proposed in \citet{Brook:2011} with an
extension in~\citep{Kaijen:ICRAWS:2011}. Different
grasp planners provide grasp hypotheses which are integrated to reach a consensus on how to grasp a
segmented point cloud. The authors show results using the planner
presented in~\citep{Hsiao:10} for unknown objects in combination with
grasp hypotheses generated through fitting known objects to point
cloud clusters as described
in~\citep{Matei:2010}. Fig.~\ref{fig:familiar:Brook} shows the grasp
hypotheses for a segmented point cloud based on the input from these different
planners. 
Another example for a hybrid approach is the work
by \citet{marton11ijrr}. A set of very simple shape primitives like
boxes, cylinders and more general rotational objects are
considered. They are reconstructed from segmented point clouds by
analysis of their footprints. Parameters such as circle radius and the
side lengths of rectangles are varied; curve parameters are estimated
to reconstruct more complex rotationally symmetric objects. Given
these reconstructions, a look-up is made in a database of already
encountered objects for re-using successful grasp hypotheses. In case
no similar object is found, new grasp hypotheses are generated using
the technique presented in~\citep{Marton10IROS}. For object hypotheses
that cannot be represented by the simple shape primitives mentioned
above, a surface is reconstructed through triangulation. Grasp
hypotheses are generated using the planner presented
in~\citep{Hsiao:10}.

\section{Discussion and Conclusion}
We have identified four major areas that form open problems in the area
of robotic grasping: 

\paragraph*{Object Segmentation}
Many of the approaches that are mentioned in this survey usually
assume that the object to be grasped is already segmented from the
background. Since segmentation is a very hard problem in itself, many
methods make the simplifying
assumption that objects are standing on a planar surface.  Detecting
this surface in a 3D point cloud and performing Euclidean clustering results in a set
of segmented point clouds that serve as object
hypotheses~\citep{Rusu09ICCV-WS}. Although the dominant surface
assumption is viable in certain scenarios and to shortcut the problem
of segmentation, we believe that we need a more general approach to
solve this.

First of all, some objects might usually occur in a specific spatial
context. This can be on a planar surface, but it might also be on a
shelf or in the fridge. \citet{aydemir2012_3Dcontext} propose to learn
this context for each known object to guide the search for them. One
could also imagine that this context could help segmenting foreground
from background. 
Furthermore, there are model-based object detection
methods \citep{Azad2007,Glover:08,detry2009Pose,Collet_Romea_2011,Papazov:2012}
that can
segment a scene as a by-product of 
detection and without making strong assumptions about the environment.  
In case of unknown objects, some methods have been proposed that 
employ the interaction capabilities of a robot,
e.g. visual fixation or pushing movements with the robot hand, to
segment the
scene~\citep{MettaF03,ObjectBirth09,Kenney:2009,Niklas:IROS:11,Schiebener2011}. A
general solution towards object segmentation might be a combination of
these two methods. The robot first interacts with objects to acquire a
model. Once it has an object model, it can be used for detecting and
thereby segmenting it from the background.

\paragraph*{Learning to Grasp}
Let us consider the goal of having a robotic companion
helping us in our household. In this scenario, we cannot expect that the
programmer has foreseen all the different situations that this robot
will be confronted with. 
Therefore, the ideal household robot should have the ability to continuously learn
about new objects and how to manipulate them while it is operating in
the environment. We will also not be able to rely on
having 3D models readily available of all objects the robot could
possibly encounter.
This requires the ability to learn a model that could generalize
from previous experience to new situations. Many open questions arise: How
is the experience regarding one object and grasp
represented in memory? How can success and failure be autonomously
quantified? How can a model be learned from this experience that would
generalize to new situations? Should it be a discriminative, a
generative or exemplar-based model? What are the features that encode
object affordances? Can these be autonomously learned?  In which space are we comparing
new objects to already encountered ones? Can we bootstrap learning by
using simulation or by human demonstration? The methods that we have
discussed in Section~\ref{sec:familiar} about grasping familiar
objects approach these questions. However, we are still far from a
method that answers all of them in a satisfying way.

\paragraph*{Autonomous Manipulation Planning}
Recently, more complex scenarios
than just grasping from a table top have been approached by a number of
research labs. How a robot can autonomously sequence
a set of actions to perform such a task is still an open
problem. Towards this end, \citet{tenorth12roboearth} propose a cloud
robotics infrastructure under which robots can share their experience
such as  action recipes and manipulation
strategies. An inference engine is provided for checking
whether all requirements are fulfilled for performing a full
manipulation strategy. It would be interesting to study how the
uncertainty in perception and execution can be dealt with in
conjunction with such a symbolic reasoning engine. 

When considering a complex action, grasp synthesis cannot be
considered as an isolated problem. On the contrary, higher-level
tasks influence what the best grasp in a specific scenario
might be, e.g. when grasping a specific tool. Task constraints have
not yet been considered extensively in the community. Current
approaches, e.g.~\citep{SongEHK11,DangA12}, achieve impressive results. As an
open question stands how to scale to life-long learning.

\paragraph*{Robust Execution}
It has been noticed by many researchers that inferring a grasp for a
given object is necessary but not sufficient. Only if execution is
robust to uncertainties in sensing and actuation, a grasp can succeed
with high probability. There are a number of approaches that use
constant tactile or visual feedback during grasp execution to adapt to unforeseen
situations \citep{Felip2009,Hsiao:10,PastorIROS2011,bekiroglu2011c,Hudson2012,Kazemi_2012,xavi:ifac:2011}. Tactile
feedback can be from haptic or force-torque sensors. Visual feedback
can be the result from tracking the hand and object simultaneously. 
Also in this area, there are a number of open questions. How can
tactile feedback be
interpreted to choose an appropriate corrective action
independent of the object, the task and environment? How can visual
and tactile information be fused in the controller?

\subsection{Final Notes}
In this survey, we reviewed work on data-driven grasp synthesis and
propose a categorization of the published work.  We focus on the type and level of 
prior knowledge
used in the proposed approaches and on the 
assumptions that are commonly made about the objects being manipulated.
 We identified recent trends in the field and provided a
discussion about the remaining challenges.

An important issue is the current lack of general
benchmarks and performance metrics suitable for comparing the
different approaches.  Although
various object-grasp databases are already available e.g. the Columbia
Grasp database~\citep{Goldfeder:2009}, the VisGraB
data set~\citep{VisGraB} or the playpen data set~\citep{Playpen} they
are not commonly used for comparison.  We acknowledge that one of
the reasons is that grasping in itself is highly dependent on the employed
sensing and manipulation hardware. 
There have also been robotic challenges
organized such as the DARPA Arm project~\citep{ARM} or
RoboCup@Home~\citep{RoboCup} and a framework for benchmarking has been
proposed by~\citet{Benchmarking}.  However, none of these successfully integrate all the 
subproblems relevant for benchmarking different grasping approaches. 

Given that data-driven grasp synthesis is an active field of research
and lots of work has been reported in the area, we set up a web page
that contains all the references in this survey at
\url{www.robotic-grasping.com}. They are structured according to the proposed
classification and tagged with the mentioned aspect.
The web page will be constantly updated with the most recent approaches.

\ifCLASSOPTIONcaptionsoff
  \newpage
\fi



\bibliographystyle{IEEEtranN}
{\footnotesize
\bibliography{survey}
}
%



%

\begin{IEEEbiography}[{\includegraphics[width=1in,height=1.25in,clip,keepaspectratio]{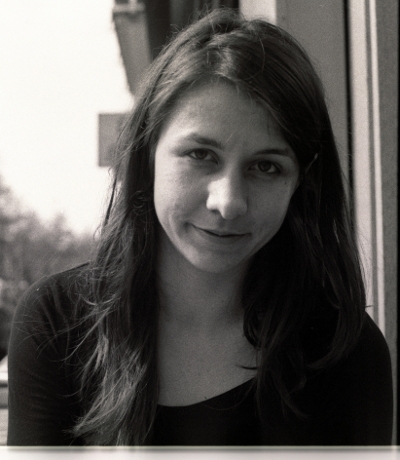}}]{Jeannette Bohg}
is a research scientist at the Autonomous Motion Department, 
Max-Planck-Institute for Intelligent Systems in T{\"u}bingen, Germany. She 
holds a Diploma in Computer Science from the
Technical University Dresden, Germany and a M.Sc. in Applied
Information Technology
from Chalmers in G{\"o}teborg, Sweden. In 2011, she received her 
PhD from the Royal Institute of Technology (KTH) in Stockholm, Sweden.   
Her research interest lies at the intersection between robotic grasping and 
Computer Vision. Specifically, she is interested in the integration of 
multiple sensor modalities and information sources for enhanced 
scene understanding. She demonstrated how this work can be used in an active 
perception framework and leads to improved grasping and manipulation.
\end{IEEEbiography}

\begin{IEEEbiography}[{\includegraphics[width=1in,height=1.25in,clip,keepaspectratio]{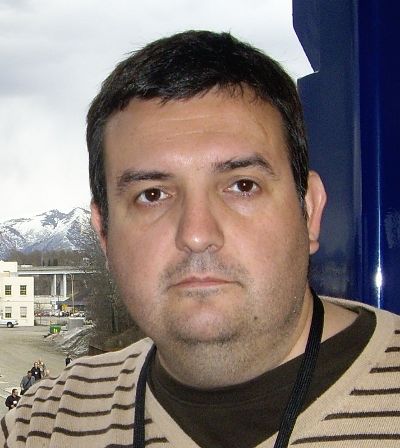}}]{Antonio Morales}
is Associate Professor at the Department of Computer Engineering and
Science in the Universitat Jaume I of Castell{\'o}, Spain. He received his
PhD in Computer Science Engineering from Universitat Jaume I
in January 2004. He is a leading researcher at the Robotic Intelligence Laboratory at
Universitat Jaume I and his research interests are focused on reactive
robot grasping and manipulation, and on development of robot
simulation. He has been a Principal Investigator of the European Cognitive
Systems Integrated Project GRASP and on several national and locally
funded research projects.

He has served as Associated Editor for the IEEE International
Conference on Robotics and Automation and for the IEEE/RSJ
International Conference on Intelligent Robots and Systems. He has
also served as reviewer for multiple relevant journals and
conferences. He is member of the IEEE-RAS society since 1998.
\end{IEEEbiography}

\begin{IEEEbiography}[{\includegraphics[width=1in,height=1.25in,clip,keepaspectratio]{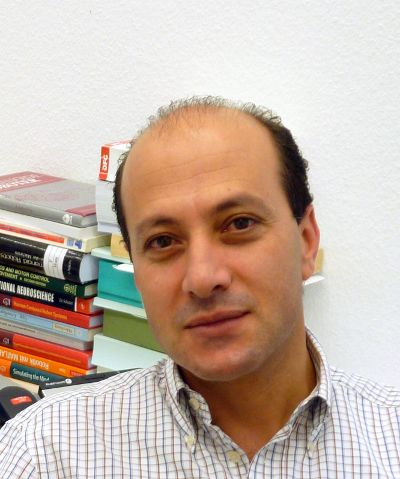}}]{Tamim Asfour}
is Professor at the Institute for Anthropomatics, Karlsruhe
Institute of Technology
(KIT). He received his diploma degree in Electrical Engineering and his PhD
in Computer Science from the University of Karlsruhe (TH).  He is developer
and leader of the development team of the ARMAR humanoid robot family. He is
European Chair of the IEEE RAS Technical Committee on Humanoid Robots and
member the Executive Board of the German Robotics Association (DGR: Deutsche
Gesellschaft für Robotik). His research interest is humanoid robotics.
Specifically, he has been researching the engineering of high performance
24/7 humanoid robots able to predict, act and interact in the real world.
His research focuses on humanoid mechatronics and mechano-informatics,
grasping and dexterous manipulation, action learning from human observation
and goal-directed imitation learning, active vision and active touch,
whole-body motion planning, robot software and hardware control architecture
and system integration.
\end{IEEEbiography}

\begin{IEEEbiography}[{\includegraphics[width=1in,height=1.25in,clip,keepaspectratio]{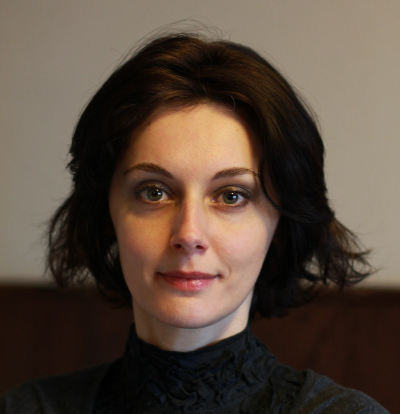}}]{Danica Kragic}
is a Professor at the School of Computer Science and 
Communication at KTH in Stockholm. She received MSc in Mechanical
Engineering from the Technical University of Rijeka, Croatia in 1995 
and PhD in Computer Science from KTH in 2001. Danica received the 
2007 IEEE Robotics and  Automation Society Early 
Academic Career Award. She is a member of the Swedish Royal Academy 
of Sciences and Swedish Young Academy. She has chaired the IEEE RAS 
Technical Committee on Computer and Robot Vision and from 2009 serves 
as an IEEE RAS AdCom member. Her research is in the area of computer vision, object 
grasping and manipulation and human-robot interaction.
Her recent work explores different learning methods for formalizing 
the models for integrated representation of objects and actions that 
can be applied on them. This work has demonstrated how robots can achieve 
scene understanding through active exploration and how full body 
tracking of humans can be made more efficient.
\end{IEEEbiography}






\end{document}